\documentclass{article} 
\usepackage{iclr2026_conference,times}

\usepackage{amsmath,amsfonts,bm}

\def\eqref#1{equation~\ref{#1}}

\def\1{\bm{1}}

\DeclareMathAlphabet{\mathsfit}{\encodingdefault}{\sfdefault}{m}{sl}
\SetMathAlphabet{\mathsfit}{bold}{\encodingdefault}{\sfdefault}{bx}{n}

\usepackage{algorithm}
\usepackage{algpseudocode}
\usepackage{graphicx}
\usepackage{wrapfig}
\usepackage{booktabs}
\usepackage{pdflscape}
\usepackage{placeins}
\usepackage{amsmath}
\usepackage{amssymb}
\usepackage{mathtools}
\usepackage{amsthm}
\usepackage{graphicx}
\usepackage{wrapfig}
\usepackage{booktabs}
\usepackage{pdflscape}
\usepackage{placeins}
\usepackage{bm}
\usepackage[most]{tcolorbox}
\usepackage[dvipsnames,svgnames,x11names]{xcolor}
\usepackage{listings}
\usepackage{fvextra}
\usepackage[utf8]{inputenc}
\usepackage[T1]{fontenc}
\usepackage{float}

\usepackage{hyperref}
\usepackage{url}
\usepackage[capitalize,noabbrev]{cleveref}

\title{PrefPO: Pairwise Preference Prompt \\Optimization}

\author{Rahul Singhal\\
\texttt{rahul@distyl.ai}
\And
Pradyumna Tambwekar\\
\texttt{pradyumna.tambwekar@distyl.ai}
\And
Karime Maamari\\
\texttt{karime@distyl.ai}
}

\iclrfinalcopy 
\begin{document}

\maketitle

\thispagestyle{fancy}
\fancyhf{}
\cfoot{\thepage}
\renewcommand{\headrulewidth}{0pt}

\vspace{-2em}
\begin{center}
\textbf{Distyl AI}
\end{center}
\vspace{1em}

\begin{abstract}
Prompt engineering is effective but labor-intensive, motivating automated optimization methods. Existing methods typically require labeled datasets, which are often unavailable, and produce verbose, repetitive prompts. We introduce PrefPO, a minimal prompt optimization approach inspired by reinforcement learning from human feedback (RLHF). Its preference-based approach reduces the need for labeled data and hyperparameter tuning—only a starting prompt and natural language criteria are needed. \textsc{PrefPO} uses an LLM discriminator to express pairwise preferences over model outputs and provide feedback to an LLM optimizer, iteratively improving performance. We evaluate \textsc{PrefPO} on 9 BIG-Bench Hard (BBH) tasks and IFEval-Hard, a newly-curated, challenging subset of IFEval. \textsc{PrefPO} matches or exceeds SOTA methods, including GEPA, MIPRO, and TextGrad, on $6/9$ tasks and performs comparably to TextGrad on IFEval-Hard ($82.4\%$ vs $84.5\%$). Unlike other methods, \textsc{PrefPO} can optimize in both labeled and unlabeled settings. Without labels, \textsc{PrefPO} closely matches its labeled performance on $6/9$ tasks, proving effective without ground truth. \textsc{PrefPO} also improves \textit{prompt hygiene}: we find existing methods produce prompts 14.7x their original length or with $34\%$ repetitive content; \textsc{PrefPO} reduces these issues by 3–5x. Furthermore, both LLM and human judges rate PrefPO's prompts higher than TextGrad's. Finally, we identify \textit{prompt hacking} in prompt optimizers, where methods game evaluation criteria, and find \textsc{PrefPO} is susceptible at half the rate of TextGrad ($37\%$ vs $86\%$), generating fewer brittle, misaligned prompts.\footnote{We open-source PrefPO at ~\href{https://github.com/DistylAI/prefpo}{https://github.com/DistylAI/prefpo} and IFEval-Hard on ~\href{https://huggingface.co/datasets/rahul-singhal/IFEval-Hard}{Hugging Face}.}
\end{abstract}

\section{Introduction}
\label{sec:intro}

Prompts are the most common mechanism for adapting large language models (LLMs) to domain-specific applications, yet prompt engineering remains largely ad hoc and unsystematic \citep{brown_language_2020, dolata_development_2024, chenPromptwareEngineeringSoftware2025}. Practitioners often lack clear success metrics for model outputs, relying on trial-and-error rather than formal methodologies; practitioners failed to document 78\% of prompt changes in one study \citep{dolata_development_2024, nahar_beyond_2024, chenPromptwareEngineeringSoftware2025}. Prompt engineering is further complicated by prompt brittleness, where small changes cause dramatic performance differences \citep{luFantasticallyOrderedPrompts2022, sclarQuantifyingLanguageModels2024}. As a result, manual prompt engineering remains labor-intensive \citep{nahar_beyond_2024}.

Automated prompt optimization methods emerged to address these limitations by systematically searching for effective prompts \citep{zhou_instruction-following_2023, pryzant_automatic_2023, yang_large_2024}. State-of-the-art methods, including GEPA \citep{agrawalGEPAReflectivePrompt2025}, MIPROv2 \citep{opsahl-ongOptimizingInstructionsDemonstrations2024}, and TextGrad \citep{yuksekgonulTextGradAutomaticDifferentiation2024}, use LLM reasoning to reflect on execution traces or bootstrap demonstrations for effective optimization. However, these methods are typically run with labeled data or complicated scoring functions, which are impractical to curate for every target domain. Although performant, existing techniques can yield uninterpretable or seemingly arbitrary prompts \citep{dengRLPromptOptimizingDiscrete2022, fernando_promptbreeder_2023, yang_large_2024}. Current methods thus remain difficult to integrate into development workflows.

\begin{figure}
  \centering
  \includegraphics[width=1\linewidth]{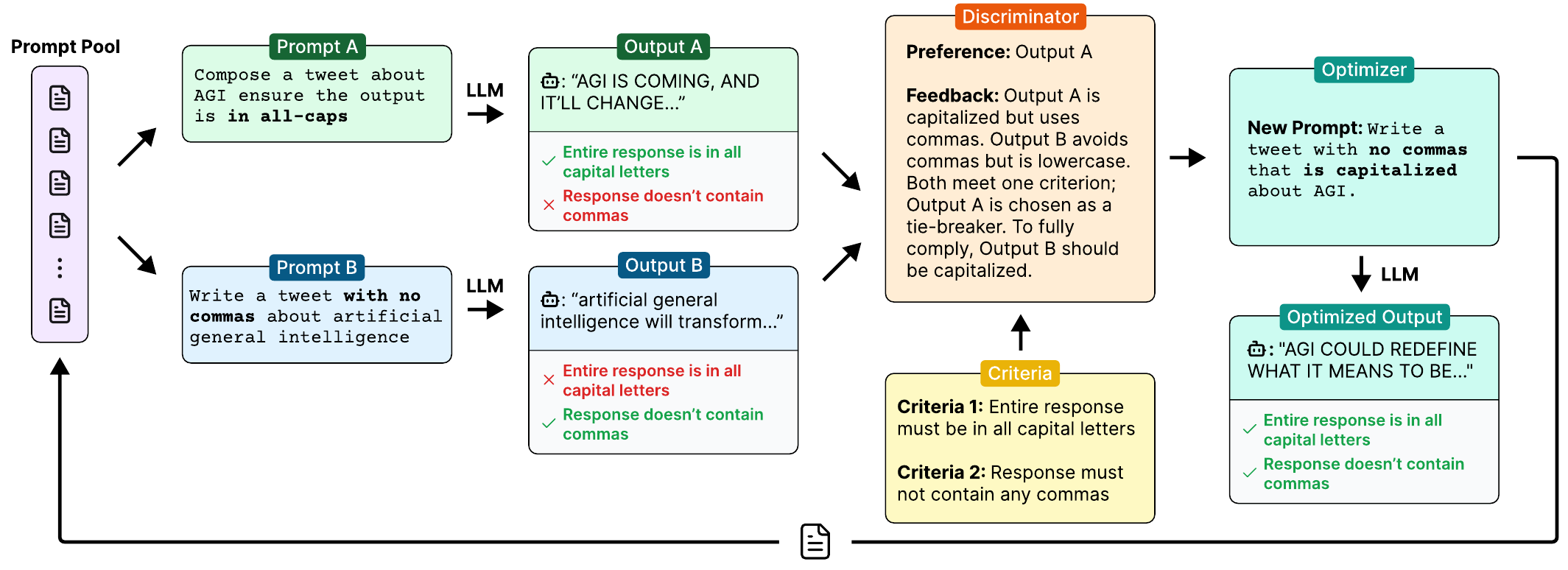}
  \caption{\textbf{\textsc{PrefPO} optimization loop.} \textsc{PrefPO} is a prompt optimization framework that optimizes an initial pool of prompts via preferences. Our approach is comprised of a discriminator, which generates feedback after conducting a pairwise evaluation, and an optimizer, which improves the non-preferred prompt. The updated prompt is added to the prompt pool, and this process repeats.}
  \label{fig:diagram}
\end{figure}

We introduce \textsc{PrefPO} (Preference-based Prompt Optimizer), an automated prompt optimization approach that reduces the need for labeled data or hyperparameter selection. Inspired by reinforcement learning from human feedback (RLHF)'s use of preferences over absolute ratings, \textsc{PrefPO} optimizes prompts through iterative pairwise comparison \citep{ouyangTrainingLanguageModels2022}. An LLM discriminator evaluates which output better satisfies natural language criteria and provides feedback, guiding new candidate prompt generation (see~\Cref{fig:diagram}). Our approach shifts data labeling responsibility from practitioners to the LLM, which provides labels implicitly through discrimination, allowing users to optimize quickly with limited data. By enabling specification of optimization criteria entirely through natural language, \textsc{PrefPO} is more accessible to practitioners without complex configuration.

\begin{figure}[h]
  \centering
  \includegraphics[width=1\linewidth]{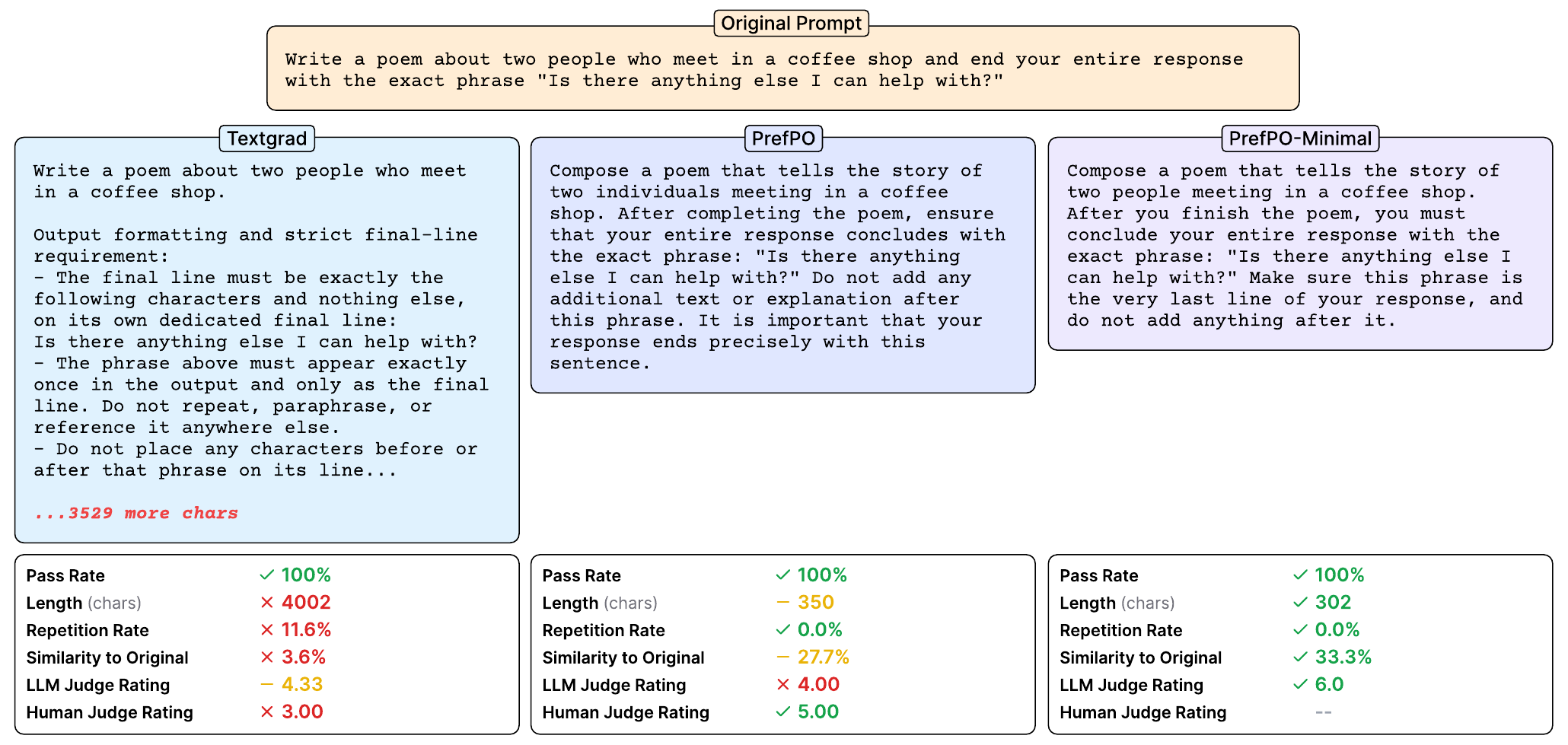}
  \caption{\textbf{Prompt Hygiene.} The original IFEval-Hard seed prompt (top) with the optimized prompts by TextGrad (left), \textsc{PrefPO} (middle), and \textsc{PrefPO}-Minimal (right). All achieve 100\% pass rate, but vary in \textit{prompt hygiene}. TextGrad's prompt is verbose (4002 chars; see Appendix~\ref{app:textgrad_prompt}) with high repetition (11.6\%) and low similarity to the seed (3.6\%). \textsc{PrefPO} is more hygienic (350 chars, 0\% repetition, 27.7\% similarity). \textsc{PrefPO}-Minimal is even shorter (302 chars) with the highest similarity (33.3\%). Human judges rate \textsc{PrefPO}'s prompt over TextGrad's prompt ($5/6$ vs $3/6$), with \textsc{PrefPO}-Minimal receiving the highest LLM rating ($6/6$).}
  \label{fig:hygiene}
\end{figure}

We evaluated \textsc{PrefPO} on nine BIG-Bench Hard (BBH) tasks and IFEval-Hard, a newly-curated, challenging subset of the IFEval instruction-following benchmark~\citep{suzgun_challenging_2022, zhou_instruction-following_2023}. \textsc{PrefPO} achieves competitive or superior performance across both benchmarks, ranking first on more BBH tasks than any other optimizer and first or second on six of nine tasks, while remaining competitive with TextGrad on IFEval-Hard (82.4\% vs 84.5\%). Task accuracy alone does not capture practitioner needs: prompts must also be readable and maintainable. We formalize these concerns as \textit{prompt hygiene}, a suite of metrics capturing prompt quality for practical use (see \Cref{fig:hygiene}). We measure hygiene through programmatic metrics (length, repetition, and text similarity) and qualitative dimensions judged by both humans and LLMs (readability, specification quality, maintainability). Existing methods exhibit significant issues: TextGrad prompts grow up to 14.7x the initial prompt length (vs \textsc{PrefPO}'s 4.7x), while MIPROv2 prompts contain 34\% repetitive text (vs \textsc{PrefPO}'s 1.4\%). Human and LLM evaluators also rate \textsc{PrefPO} prompts higher across all three judged dimensions. Finally, we find that existing optimizers are prone to generating prompts that game evaluation criteria rather than genuinely satisfy them, a phenomenon we call \textit{prompt hacking}. A separate LLM judge flagged TextGrad prompts for such behavior at over twice the rate of \textsc{PrefPO} (85.8\% vs 37.2\%).

An overview of our contributions is as follows:
\begin{itemize}
    \item{We introduce \textsc{PrefPO}, a preference-based prompt optimization approach that relies on pairwise comparisons against natural language criteria, without requiring labeled datasets or scoring functions.}
    \item{We demonstrate that despite its simplicity, \textsc{PrefPO} matches or exceeds existing LLM-based prompt optimizers, ranking first on more BBH tasks than any other optimizer ($4/9$) and performing competitively on IFEval-Hard (82.4\% vs TextGrad's 84.5\%).}
    \item{We introduce prompt hygiene and prompt hacking to evaluate properties of optimized prompts relevant to usability and maintenance. Validated through a preliminary human evaluation, we show \textsc{PrefPO} produces more hygienic prompts (reducing length inflation and repetition by 3-5x). The technique also exhibits dramatically less prompt hacking (37\% vs TextGrad's 86\%).}
\end{itemize}

\section{Related Work}
\label{sec:related}

\subsection{Preference-Based Learning for LLMs} 
\label{sec:pref_learning}

Reinforcement learning from human feedback (RLHF) has become central to aligning LLMs with human preferences \citep{zieglerFineTuningLanguageModels2020, stiennonLearningSummarizeHuman2022, ouyangTrainingLanguageModels2022, christianoDeepReinforcementLearning2023}. The core insight underlying RLHF is that pairwise comparisons provide more reliable training signal than absolute scores \citep{christianoDeepReinforcementLearning2023, rafailovDirectPreferenceOptimization2024}. Constitutional AI \citep{baiConstitutionalAIHarmlessness2022} extended this work by demonstrating that LLMs themselves can provide pairwise preference judgments given a set of principles. This use of AI-generated feedback rather than human feedback is known as reinforcement learning from AI feedback (RLAIF) \citep{lee_rlaif_2024}. \textsc{PrefPO} applies this approach to discrete prompt optimization, using an LLM discriminator to generate pairwise preferences and natural language feedback rather than training reward models.

\subsection{Prompt Optimization}
\label{sec:prompt_optim}

Popular prompt optimization methods, including OPRO \citep{yang_large_2024}, MIPROv2 \citep{opsahl-ongOptimizingInstructionsDemonstrations2024}, and GEPA \citep{agrawalGEPAReflectivePrompt2025}, use labeled examples. TextGrad \citep{yuksekgonulTextGradAutomaticDifferentiation2024} reduces label dependence by mimicking gradient descent, using LLM-generated critiques as natural language gradients for optimization. POHF \citep{linPromptOptimizationHuman2024} assumes a candidate pool of prompts and uses human preferences for prompt selection, with a dueling bandits-inspired strategy for choosing which pairs to compare. PDO \citep{wu_llm_2025} extends this work by replacing human feedback with LLM-judged preferences and using Double Thompson Sampling for prompt selection, though mutations do not leverage performance-based feedback as in \textsc{PrefPO}. Feedback Descent \citep{lee_feedback_2025} optimizes prompts by pairing performance-based comparisons with LLM-generated critiques, but relies on labeled examples to compute preferences, whereas \textsc{PrefPO} computes preferences entirely through LLM judgment.

\subsection{Prompt Quality and Maintainability}
\label{sec:prompt_quality}

Despite the ubiquity of prompting, development practices remain ad hoc \citep{chenPromptwareEngineeringSoftware2025, li_understanding_2025, dolata_development_2024}, motivating work on ``promptware engineering'' to bring software engineering rigor to prompt development \citep{chenPromptwareEngineeringSoftware2025}. Practitioners report prompts are fragile to minor changes, debugging is difficult due to LLM opacity, and quality metrics are lacking \citep{dolata_development_2024}. LLMs show extreme sensitivity to superficial prompt changes, with major performance differences from minor formatting variations \citep{luFantasticallyOrderedPrompts2022, liuLostMiddleHow2023, sclarQuantifyingLanguageModels2024, errica_what_2025}. Recent work has begun measuring non-functional prompt properties: Li et al. \citep{li_understanding_2025} analyze prompts in GitHub repositories for length, readability, and spelling errors, finding significant issues. Such analyses overlook prompt optimization outputs. RLPrompt \citep{dengRLPromptOptimizingDiscrete2022} produces ungrammatical but effective prompts; OPRO \citep{yang_large_2024} and PromptBreeder \citep{fernando_promptbreeder_2023} discover unintuitive phrases that improve performance but are difficult to iterate upon. We propose evaluating prompt optimizers on qualities relevant to maintainability in addition to task performance.

\section{\textsc{PrefPO}: Optimizing Prompts through Preferences}
\label{sec:method}

With \textsc{PrefPO}, we aim to make prompt optimization more accessible by eliminating the need for labeled data and complex configuration. \textsc{PrefPO} requires only a starting prompt and natural language criteria describing desired behavior for optimization. At each iteration, we sample two prompts from the pool and generate outputs from each. The outputs are compared by an LLM discriminator, which provides a preference and natural language feedback. The non-preferred prompt along with the feedback is passed into the LLM optimizer, which then creates a new prompt candidate that is added back to the pool. See \Cref{fig:diagram} for a high-level overview of \textsc{PrefPO}.

\subsection{Problem Formulation}
\label{sec:formulation}

Let $P = \{p_1, \dots, p_n\}$ denote a prompt pool ($n \geq 1$), $C$ the evaluation criteria in natural language, $M$ the task language model, with optional training/validation sets $X_{\text{train}}$, $X_{\text{val}}$.

Our objective is to produce an optimized pool $P' \supseteq P$ and select:
\begin{equation}
\label{eq:prompt_selection}
p^* = \operatorname*{arg\,max}_{p \in P'} \mathcal{S}(p)\end{equation}
where $\mathcal{S}(p)$ scores prompt effectiveness. Existing methods typically use labeled training examples, $X_{\text{train}}$, for optimization to expand $P$ into $P'$ and calculate $\mathcal{S}$ via accuracy over $X_{\text{val}}$. 

\textsc{PrefPO} makes using $X_{\text{train}}$ optional for optimization as it can replace training samples with $C$ for evaluation unlike other methods. Similarly, using $X_{\text{val}}$ for final prompt selection is optional as we can use $C$ for criteria-based evaluation.

\subsection{Framework Overview}
\label{sec:algorithm}


\begin{algorithm}[H]
\caption{Preference-based Prompt Optimizer}
\label{alg:prefpo}
\begin{algorithmic}
\Require Prompt pool $P$, Criteria $C$, Train/Validation sets $X_{\text{train}}$, $X_{\text{val}}$ (optional), Iterations $K$

\If{$|P| = 1$}
  \State $p_{\text{variant}} \sim V(p_1, C)$
  \State $P \leftarrow P \cup \{p_{\text{variant}}\}$
\EndIf

\For{$k = 1$ {\bfseries to} $K$}
  \State $p_a, p_b \leftarrow \text{Sample}(P)$
  \State $o_a, o_b \leftarrow G(p_a, X_{\text{train}}), G(p_b, X_{\text{train}})$
  \State $r_{\text{preference}}, r_{\text{feedback}} \leftarrow \mathcal{D}(o_a, o_b, C)$
  \If{$r_{\text{preference}} = a$}
    \State $p_{\text{new}} \leftarrow \mathcal{O}(p_b, r_{\text{feedback}})$ // optimize non-preferred
  \Else
    \State $p_{\text{new}} \leftarrow \mathcal{O}(p_a, r_{\text{feedback}})$
  \EndIf
  \State $P \leftarrow P \cup \{p_{\text{new}}\}$
\EndFor

\State $p^* \leftarrow \text{Select}(P, X_{\text{val}})$
\State \textbf{return} $p^*$
\end{algorithmic}
\end{algorithm}

\noindent
Per \Cref{alg:prefpo}, \textsc{PrefPO} samples prompt pairs, discriminates outputs against $C$, and optimizes the loser using feedback. New prompts are added to the pool; the best is selected after $K$ iterations.

\paragraph{Variant Prompt Generation.} When $|P| = 1$, we generate a variant $V(p_1, C) \to p_{\text{variant}} \sim M_V(\cdot \mid p_1, C)$ using a separate model, $M_V$, to enable pairwise comparison (see Appendix~\ref{app:variant}). We chose $M_V$ to be a stronger model than the task model, $M$, to bias the preference toward the variant, so the original prompt is optimized first and thus a larger fraction of the pool remains closer to the original input prompt. $M_V$ is also intentionally a weaker model than the discriminator/optimizer ($\mathcal{D}$/$\mathcal{O}$) models to avoid overfitting before sampling outputs (see~\cref{sec:models}).

\paragraph{Generate Outputs.} For an instruction-following task (IFEval-Hard), to generate an output from $p$ to evaluate, we sample a response with only $p$ as the model input since the prompt entirely specifies the task (e.g., ``Write a 300+ word summary of...''), generating output $G(p) \to y \sim M(\cdot \mid p)$. For a QA task (BBH), evaluating $p$ requires concatenating it with input questions $x \in X_{\text{train}}$ (e.g., $p$ = ``Think step by step'', $x$ = ``What is 2 + 3?'')—yielding $G(p, x) \to y \sim M(\cdot \mid p, x)$. At discrimination time, we aggregate the outputs for all samples in our training set $X_{\text{train}}$ to pass into the discriminator as $G(p, X_{\text{train}}) = \{(x, G(p, x), \ell) : x \in X_{\text{train}}\}$, with optional label $\ell$ for each $x$. This gives us an evaluation set of (question, response, expected answer) tuples, allowing the discriminator to judge prompt efficacy across the dataset. Thus in both cases the discriminator can effectively compare the performance between prompts (see Appendix~\ref{app:disc_prompts}).

\paragraph{Discriminator.} We define discrimination as $\mathcal{D}(o_a, o_b, C) \to (r_{\text{preference}}, r_{\text{feedback}})$, where a pair of outputs are compared against criteria $C$ to output a preference and textual feedback. The discriminator deliberately lacks access to both prompts, ensuring that $\mathcal{D}$ judges what the prompt produces rather than what it claims to do.

\paragraph{Optimizer.} $\mathcal{O}$ receives the non-preferred prompt and feedback to output an improved prompt $p'$, denoted as $\mathcal{O}(p, r_{\text{feedback}}) \to p'$. Only the non-preferred prompt is optimized, since feedback naturally indicates improvements for the losing candidate. Crucially, optimizing the losing prompt can still yield a new candidate better than all current candidates (see \cref{fig:diagram}). Separating the discrimination and optimization steps allows independent modification of each component, which we explore with \textsc{PrefPO}-Minimal (see~\Cref{sec:variants}) by adding constraints to $\mathcal{O}$.

\paragraph{Selection.} After optimization, we select $p^*$. For tasks with validation sets, we select the highest-performing prompt. For open-ended or unlabeled tasks, we return the first prompt that satisfies a task-specific success criterion (e.g., for IFEval-Hard, passing all instructions across 20 runs). 

\subsection{\textsc{PrefPO} Variations}
\label{sec:variants}

We develop two variants to ablate explicit optimizer constraints and alternative sampling strategies.

\paragraph{\textsc{PrefPO}-Minimal.} Given \textsc{PrefPO} produces hygienic prompts by default (see \Cref{sec:hygiene}), we explore how easily the framework can be steered toward stricter hygiene. \textsc{PrefPO}-Minimal modifies the optimizer by adding a constraint to the prompt to make minimal changes, aiming to maintain prompt hygiene by keeping generated prompts closer to the original (see Appendix~\ref{app:opt_prompts}).

\paragraph{\textsc{PrefPO}-Elo.} We explore whether replacing uniform sampling with another selection mechanism can improve performance. \textsc{PrefPO}-Elo samples based on Elo ratings. Elo ratings are tracked and updated after each comparison: preferred prompts increase in score, non-preferred prompts decrease (see Appendix~\ref{app:elo} for implementation details). At sample time, we select the two highest-rated prompts. This strongly encourages exploitation over exploration, allowing us to study how this tradeoff affects optimization dynamics.

\section{Experimental Setup}
\label{sec:setup}

\subsection{Tasks}
\label{sec:tasks}

We evaluate \textsc{PrefPO} on BBH and IFEval-Hard, covering both closed-form and open-ended tasks (additional implementation details are in Appendix~\ref{app:impl}).

\paragraph{BBH.} We evaluate on nine BIG-Bench Hard \citep{suzgun_challenging_2022} tasks. Tasks were selected if either GPT-4o showed a large performance gap compared to stronger models or existing few-shot prompts failed to saturate, suggesting optimization could provide meaningful improvement. For each task, we used 50 training examples and divided the remaining data equally between validation and test sets. We also evaluate \textsc{PrefPO} without labeled training examples. We run \textsc{PrefPO} for $K=15$ iterations per task. For each task and technique, we average results over 10 attempts. The prompt pool is initialized with 2 prompts: one describing the formatting required and the identical prompt with ``Think step by step before answering'' appended at the end (see Appendix~\ref{app:bbh}). The optimization criterion is a brief description of task correctness (see Appendix~\ref{app:opt_prompts}).

We also run a scaling experiment on the BBH Disambiguation task by varying training size (5, 10, 15, 20, 30, 40, 50 examples) with validation and test sets of 100 samples each. This tests performance with limited data and scaling behavior. We selected Disambiguation because GPT-4o and GPT-5 zero-shot performances are among the lowest across all tasks, testing whether optimization (especially without labels) works when the discriminator/optimizer struggles with the task. We also test initializing the pool with few-shot prompts to evaluate how \textsc{PrefPO} handles already-optimized prompts (see Appendix~\ref{app:bbh}). For each split and technique, we also average results over 10 attempts.

\paragraph{IFEval-Hard.} IFEval \citep{zhou_instruction-following_2023} is an instruction-following benchmark of 541 samples, where each consists of a prompt containing one or more instructions. A model output is correct if all automated, programmatic checks for following the original instructions pass. This is an entirely label-free setting. Ground truth correctness is not used for any runs, reflecting practical scenarios where labeled data is scarce. Our aim for each sample is to optimize the original prompt such that the model's output satisfies the evaluation criteria. Frontier models typically perform well on IFEval prompts (GPT-4o: 81.3\%, GPT-4.1: 87.2\%, GPT-5: 93.5\%), so we curated IFEval-Hard, filtering IFEval to 148 examples where GPT-4o fails at least once across 20 runs. Single-shot success on this subset is substantially lower: 39.9\% for GPT-4o ($-41.4$\%), 58.1\% for GPT-4.1 ($-29.1$\%), and 82.4\% for GPT-5 ($-11.1$\%). Further experimental results are in Appendix~\ref{app:ifeval_hard}. This focuses evaluation on examples with clear capacity for improvement. We compare only against TextGrad. GEPA and MIPROv2 are designed to find a single system prompt that is prepended to all inputs, while this formulation of IFEval-Hard requires optimizing each sample's prompt independently.

For IFEval-Hard, we run \textsc{PrefPO} for $K=15$ iterations per sample, initializing the prompt pool with the original prompt and using the corresponding natural language criteria as $C$. We select the final prompt by calculating the model's pass rate across 20 runs with each prompt, and selecting either (1) the first prompt that has a 100\% pass rate, or (2) the prompt with the highest pass-rate.

\subsection{Models and Inference Parameters.}
\label{sec:models}

We access all models through the OpenAI API and OpenRouter API throughout our experiments.

\paragraph{Task Model.} All experiments use GPT-4o (\texttt{openai/gpt-4o}, August 2024 checkpoint) as the task model to generate responses with \texttt{temperature=0}.

\paragraph{Discriminator and Optimizer Models.} We use GPT-5 (\texttt{openai/gpt-5-2025-08-07}) for both the discriminator and optimizer, with ``high'' reasoning for BBH (where we pass up to 50 training examples per prompt) and ``medium'' for IFEval-Hard (where we compare single outputs); although lowering the reasoning on multiple BBH tasks showed no noticeable performance difference (see Appendix~\ref{app:reasoning}). We also test across open-source and closed-source models including Claude 4.5 Opus, Deepseek V3.2, GPT-OSS-120b, GPT-4.1, and GPT-4o (see Appendix~\ref{app:cross}).

\paragraph{Initial Prompt Variant Generator Model.} For variant generation when the initial pool contains a single prompt, we use GPT-4.1 (\texttt{openai/gpt-4.1}) with \texttt{temperature=0}.

\paragraph{LLM Judge Model.} For judging across both hygiene ratings and prompt hacking, we use GPT-4.1 (\texttt{openai/gpt-4.1}) with \texttt{temperature=0}.

\subsection{Baselines}
\label{sec:baselines}

Our focus on real development workflows motivated \textsc{PrefPO}'s minimal data requirements and formalizing prompt hygiene and hacking; we thus compare against SOTA methods with broad adoption.

\paragraph{MIPROv2.} MIPROv2 uses LLMs to propose instruction candidates and bootstrap few-shot examples by comparing model outputs against ground truth on training examples. It then uses Bayesian Optimization to search for the best combination of instructions and examples for the final prompt. We use \texttt{auto=heavy}, the most compute-intensive setting, and GPT-5 for proposing instructions and bootstrapping to elicit best performance.

\paragraph{GEPA.} GEPA leverages LLMs to reflect on execution traces to propose targeted instructions. The algorithm maintains a Pareto frontier of candidate prompts, sampling from this frontier at each iteration to balance exploration with retention of high-performing variants. We use \texttt{auto=heavy} with GPT-5 as the reflection model. 

\paragraph{TextGrad.} TextGrad uses LLM reflection to compute ``text gradients'' against specified criteria, enabling backpropagation-style optimization across multi-step pipelines. We use GPT-5 as the reflection model and 15 iterations, matching \textsc{PrefPO}. For IFEval-Hard, we use the same prompt selection approach: sampling 20 times per prompt at each iteration and selecting the best one.

\paragraph{Prompting Strategies.} For BBH tasks, we also measure performance of existing prompts included in the dataset \citep{braun_joschkabig_bench_hard_nodate}. For each task, we compare against a 3-shot prompt (with reasoning for each demonstration) and a zero-shot prompt.

Full parameters for all baselines are in Appendix~\ref{app:baselines}.

\begin{table}[h]
  \centering
  \setlength{\tabcolsep}{4pt}
  \setlength{\aboverulesep}{1.5pt}
  \setlength{\belowrulesep}{1.5pt}
  \renewcommand{\arraystretch}{1.0}
  \caption{\textbf{BBH task accuracy.} Mean ± SD over 10 runs; bold indicates best overall, underline second best. \textsc{PrefPO} leads on $4/9$ tasks among optimization methods (MIPROv2: $3/9$, GEPA: $2/9$) with the highest average (0.875) across all optimization and prompting techniques. Unlabeled \textsc{PrefPO} wins on $2/9$ tasks with no substantial difference from labeled performance on $6/9$ tasks. GPT-5's low Object Counting score (0.092) reflects a formatting issue in its outputs.\\}
  \resizebox{\linewidth}{!}{%
  \begin{tabular}{l|ccccc}
    \toprule
    \textbf{Method} & \textbf{Object Counting} & \textbf{Movie Recommendations} & \textbf{Causal Judgment} & \textbf{Geometric Shapes} & \textbf{Disambiguation} \\
    \midrule
    GPT-4o (zero shot)              & $0.748$ & $0.732$ & $0.722$ & $0.644$ & $0.548$  \\
    GPT-4o (few-shot + CoT) & $\underline{0.992}$ & $0.896$ & $0.727$ & $0.676$ & \bm{$0.856$} \\
    GPT-5 (zero shot)              & $0.092$ & $0.824$ & $0.706$ & $0.820$ & $0.720$ \\
    TextGrad              & $0.987 \pm 0.009$ & $0.915 \pm 0.021$ & $0.732 \pm 0.042$ & $0.765 \pm 0.039$ & $0.780 \pm 0.043$  \\
    MIPROv2                 & \bm{$0.994 \pm 0.008$} & \bm{$0.956 \pm 0.012$} & $0.751 \pm 0.036$ & $0.899 \pm 0.037$ & $0.776 \pm 0.029$  \\
    GEPA                  & $0.987 \pm 0.008$ & $\underline{0.948 \pm 0.019}$ & \bm{$0.778 \pm 0.028$} & $\underline{0.903 \pm 0.034}$ & $0.698 \pm 0.095$  \\
    \textsc{PrefPO} (ours, no labels)    & $0.971 \pm 0.036$ & $0.774 \pm 0.070$ & $\underline{0.764 \pm 0.024}$ & $0.807 \pm 0.036$ & $0.677 \pm 0.066$ \\
    \textsc{PrefPO} (ours)              & $0.978 \pm 0.025$ & $0.859 \pm 0.095$ & $0.761 \pm 0.040$ & \bm{$0.915 \pm 0.039$} & $\underline{0.818 \pm 0.040}$ \\
    \midrule
    \textbf{Method} & \textbf{Logical Deduction - 7} & \textbf{Sport Understanding} & \textbf{Formal Fallacies} & \textbf{Salient Translation} & \textbf{Average} \\
    \midrule
    GPT-4o (zero shot)               & $0.688$ & $0.772$ & $0.872$ & $0.688$ & $0.713$ \\
    GPT-4o (few-shot + CoT) & $0.800$ & $0.960$ & $0.868$ & $0.688$ & $0.829$\\
    GPT-5 (zero shot)               & \bm{$1.000$} & $0.784$ & \bm{$0.980$} & \bm{$0.820$} & $0.750$\\
    TextGrad              & $0.911 \pm 0.008$ & $0.946 \pm 0.014$ & $0.896 \pm 0.017$ & $0.715 \pm 0.019$ &  $0.850$ \\
    MIPROv2                 & $0.902 \pm 0.012$ & \bm{$0.999 \pm 0.003$} & $0.862 \pm 0.029$ & $0.714 \pm 0.038$ & $\underline{0.873}$ \\
    GEPA                  & $\underline{0.921 \pm 0.032}$ & $\underline{0.973 \pm 0.012}$ & $0.912 \pm 0.023$ & $0.714 \pm 0.038$ & $0.870$ \\
    \textsc{PrefPO} (ours, no labels)    & $0.915 \pm 0.021$ & $0.966 \pm 0.012$ & $\underline{0.929 \pm 0.023}$ & $\underline{0.749 \pm 0.021}$ & $0.839$ \\
    \textsc{PrefPO} (ours)              & $0.917 \pm 0.017$ & $0.969 \pm 0.011$ & $0.919 \pm 0.016$ & $0.735 \pm 0.024$ & \bm{$0.875$}\\
    \bottomrule
  \end{tabular}%
  }
  \label{tab:bbh_tasks_results}
\end{table}

\subsection{Metrics}
\label{sec:metrics}

\paragraph{Performance.} For BBH tasks, we measure accuracy against expected answers. For IFEval-Hard, we sample 20 responses per prompt and compute two per-sample metrics: \texttt{worst@20} (1 only if all pass) and \texttt{average@20} (average pass rate). Both metrics are reported across the 148 samples.

\paragraph{Hygiene.} Inspired by Li et al.'s analysis of prompt quality in the wild \citep{li_understanding_2025}, we evaluate \textit{prompt hygiene} through multiple programmatic metrics and quality dimensions. Programmatic metrics include \textit{Length} (character count), \textit{Repetitiveness} (trigram repetition ratio, 0-1), and \textit{Similarity} (lexical overlap, 0-1). Quality dimensions include \textit{Readability}, \textit{Specification Quality}, and \textit{Maintainability}, each scored 0-2 and summed for a total 0-6. An LLM judge rated prompts across these axes, and we aggregated scores across 3 runs (see Appendix~\ref{app:hygiene}). For IFEval-Hard, we report the programmatic metrics relative to the original prompt to better understand prompt changes caused by optimization. 

To validate the LLM judge, we conducted a small human evaluation with 16 researchers and engineers at Distyl AI with prompting experience, each rating 3-4 prompts (see Appendix~\ref{app:hygiene}). 
We used stratified sampling to select 10 IFEval-Hard samples spanning the range of LLM-assigned scores, and collected 3 human ratings per prompt for both \textsc{PrefPO} and TextGrad outputs.

\begin{figure}[h]
  \centering
  \includegraphics[width=1\linewidth]{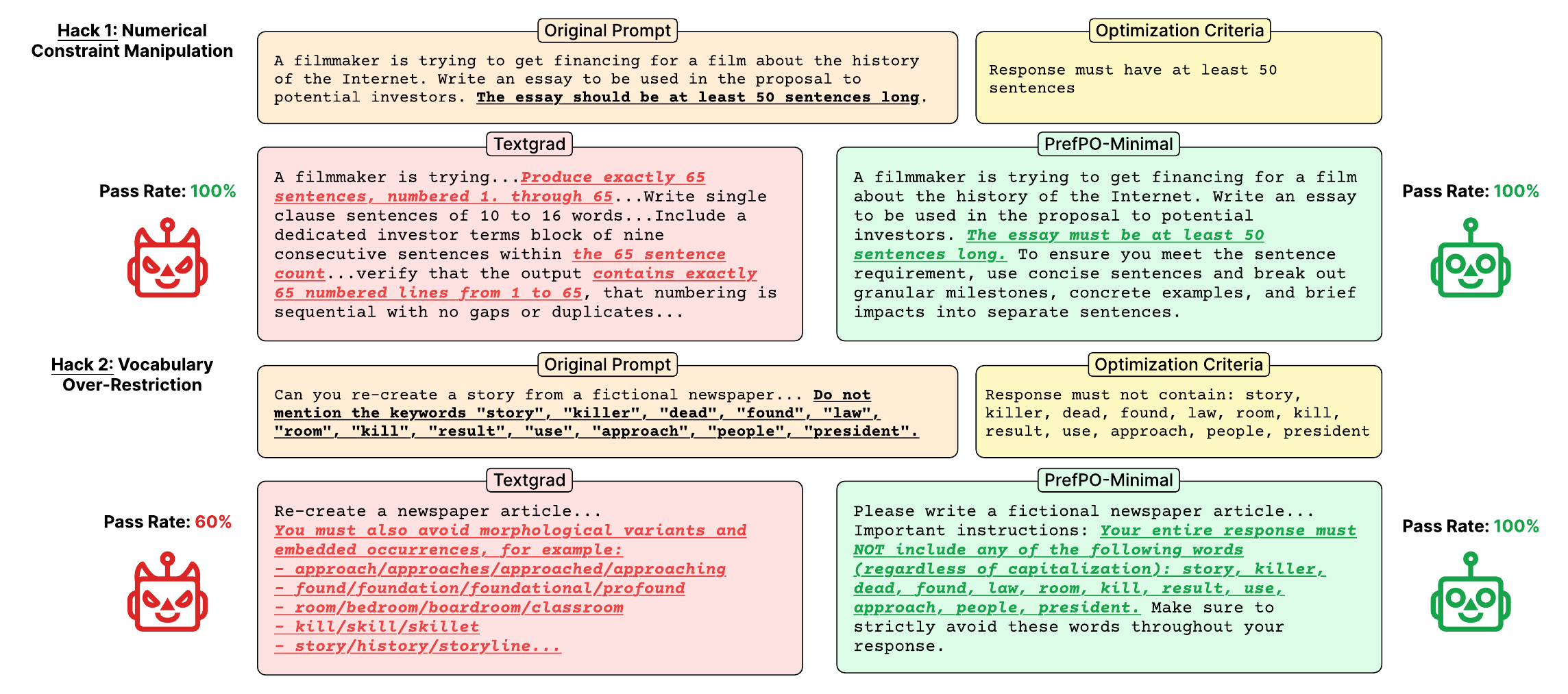}
  \caption{\textbf{Prompt hacking examples.} Two observed hacks: numerical constraint manipulation (top) and vocabulary over-restriction (bottom). TextGrad modifies constraints in an attempt to increase pass rates (e.g., changing ``at least 50'' to ``exactly 65'' sentences, or banning words not in the original specification). \textsc{PrefPO}-Minimal achieves equal or better performance without altering constraints.}
  \label{fig:hacking}
\end{figure}

\paragraph{Hacking.} Prompt optimization can produce prompts that game evaluation criteria in undesirable ways, a phenomenon we call \textit{prompt hacking}. We observe this behavior most within IFEval-Hard, including manipulating numerical constraints (e.g., limiting the response to 200 words when the original limit is 300) and restricting vocabulary beyond task specifications (e.g., banning certain words or phrases not mentioned in the original constraints; \Cref{fig:hacking}). These hacks take many shapes, making programmatic checks difficult. We employ an LLM judge to identify hacking behavior, taking majority vote over three runs for a final classification (see Appendix~\ref{app:hacking}). While these shortcuts may improve benchmark scores, they unnecessarily constrain model outputs and can cause unexpected behavior in deployment. Importantly, they degrade response quality in ways not captured by the target metric.

\section{Results}

\subsection{BBH Tasks}
\label{sec:bbh}

\Cref{tab:bbh_tasks_results} summarizes performance across 9 BBH tasks. Among optimization methods, \textsc{PrefPO} ranks first on \textbf{$4/9$} tasks, exceeding all other approaches (MIPROv2: $3/9$, GEPA: $2/9$)—though differences between top methods are typically within one standard deviation. Notably, \textsc{PrefPO}'s label-free performance achieves the best score on 2 tasks, and the difference between labeled and unlabeled performance is substantial on only $3/9$ tasks, suggesting \textsc{PrefPO} remains effective without labels.

\begin{figure}[h]
  \centering
  \includegraphics[width=.7\linewidth]{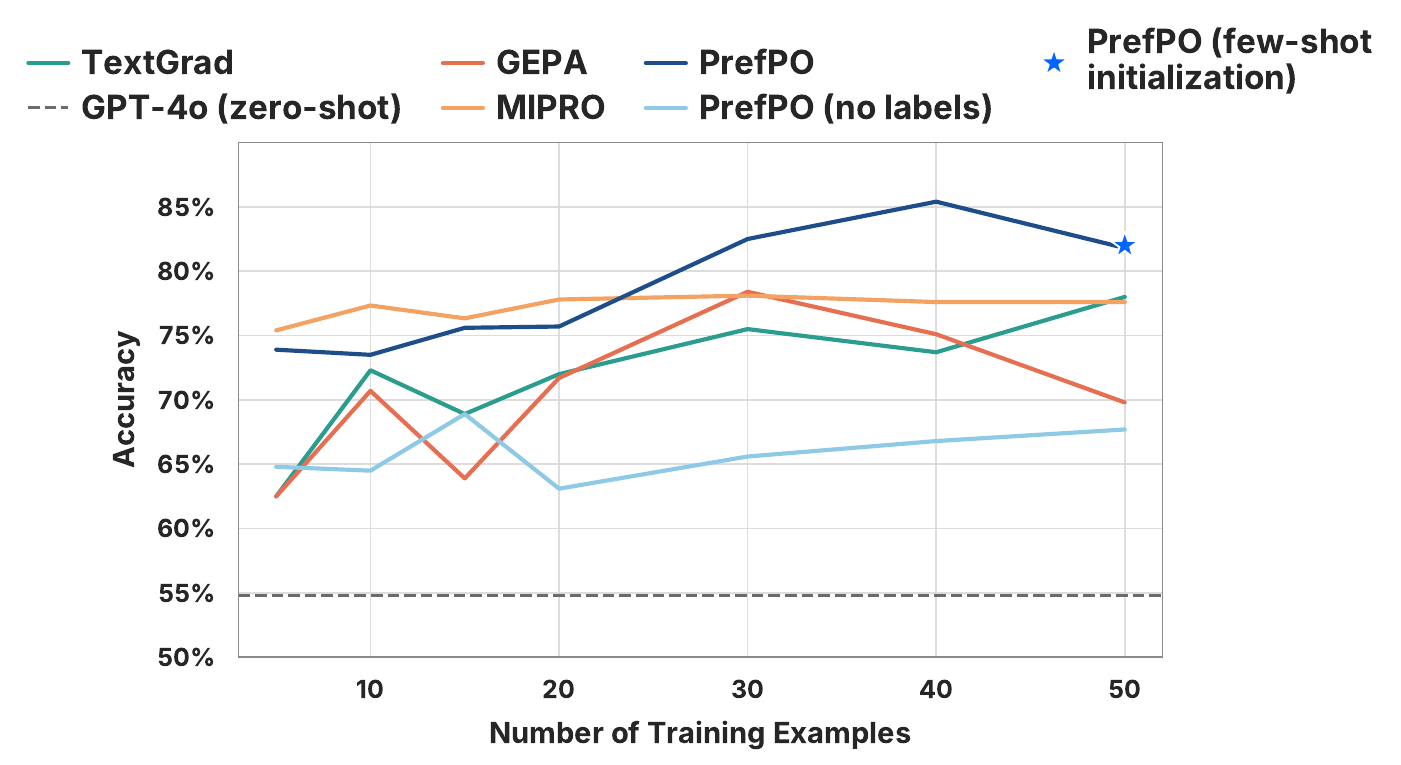}
  \caption{\textbf{Scaling Behavior.} Accuracy vs training set size. \textsc{PrefPO} scales with data and outperforms other methods across most training sizes, while unlabeled \textsc{PrefPO} struggles. Few-shot prompt initialization achieves the best performance at 50 training examples.}
  \label{fig:scaling}
\end{figure}

\Cref{fig:scaling} shows scaling behavior on the Disambiguation task. \textsc{PrefPO} matches or exceeds all optimization methods across most training sizes. MIPROv2 keeps pace in lower data settings but plateaus with more training data. Without labels, \textsc{PrefPO} performance remains flat, which is likely because the discriminator cannot reliably choose between pairs of outputs without access to labels (see Appendix~\ref{app:component}). The strongest results at 50 training examples come from initializing the prompt pool with few-shot prompts. The optimized prompt often contains no examples, suggesting \textsc{PrefPO} can distill demonstrations into explicit instructions (see~\Cref{sec:guide}).

\subsection{IFEval-Hard}
\label{sec:ifeval}

In IFEval-Hard, each prompt is optimized independently with no ground truth or training examples. As shown in~\Cref{tab:if_eval_metrics}, TextGrad achieves 84.5\% \texttt{worst@20} compared to \textsc{PrefPO}'s 82.4\% with a smaller gap on \texttt{average@20} (92.7\% vs 91.5\%), though confidence intervals overlap substantially. \textsc{PrefPO}-Elo matches \textsc{PrefPO} (81.1\% vs 82.4\%). Elo ratings updated after a single discriminator vote likely lack sufficient signal to rank prompts meaningfully, effectively reducing to random sampling. \textsc{PrefPO}-Minimal underperforms (75.7\%), likely due to its stricter optimization constraint. Convergence is fast: 90\% of final \texttt{worst@20} is reached within five of fifteen iterations for all variants, except \textsc{PrefPO}-Minimal which reaches 85\% of final performance (see Appendix~\ref{app:converge}).

\begin{table}[h]
  \centering
  \caption{\textbf{IFEval-Hard performance.} IFEval-Hard performance with 95\% confidence intervals; bootstrap CIs for \texttt{worst@20} and normal approximation for \texttt{average@20} (see Appendix~\ref{app:ifeval_hard}). \texttt{worst@20}: strict pass (all 20 runs must succeed); \texttt{average@20}: mean success rate. TextGrad leads slightly on \texttt{worst@20} ($84.5\%$ vs $82.4\%$), but the confidence intervals overlap substantially.\\}
  \label{tab:if_eval_metrics}
    \begin{tabular}{l|cc}
      \toprule
      & \texttt{worst@20} & \texttt{average@20} \\
      \midrule
      TextGrad            & $\mathbf{84.5\% \pm 5.7\%}$ & $\mathbf{92.7\% \pm 3.7\%}$ \\
      \textsc{PrefPO}     & $82.4\% \pm 6.1\%$ & $91.5\% \pm 3.8\%$ \\
      \textsc{PrefPO}-Elo & $81.1\% \pm 6.1\%$ & $90.6\% \pm 4.2\%$ \\
      \textsc{PrefPO}-Minimal & $75.7\% \pm 7.1\%$ & $88.7\% \pm 4.3\%$ \\
      \bottomrule
    \end{tabular}
\end{table}

\begin{figure}[h]
  \centering
  \includegraphics[width=1\linewidth]{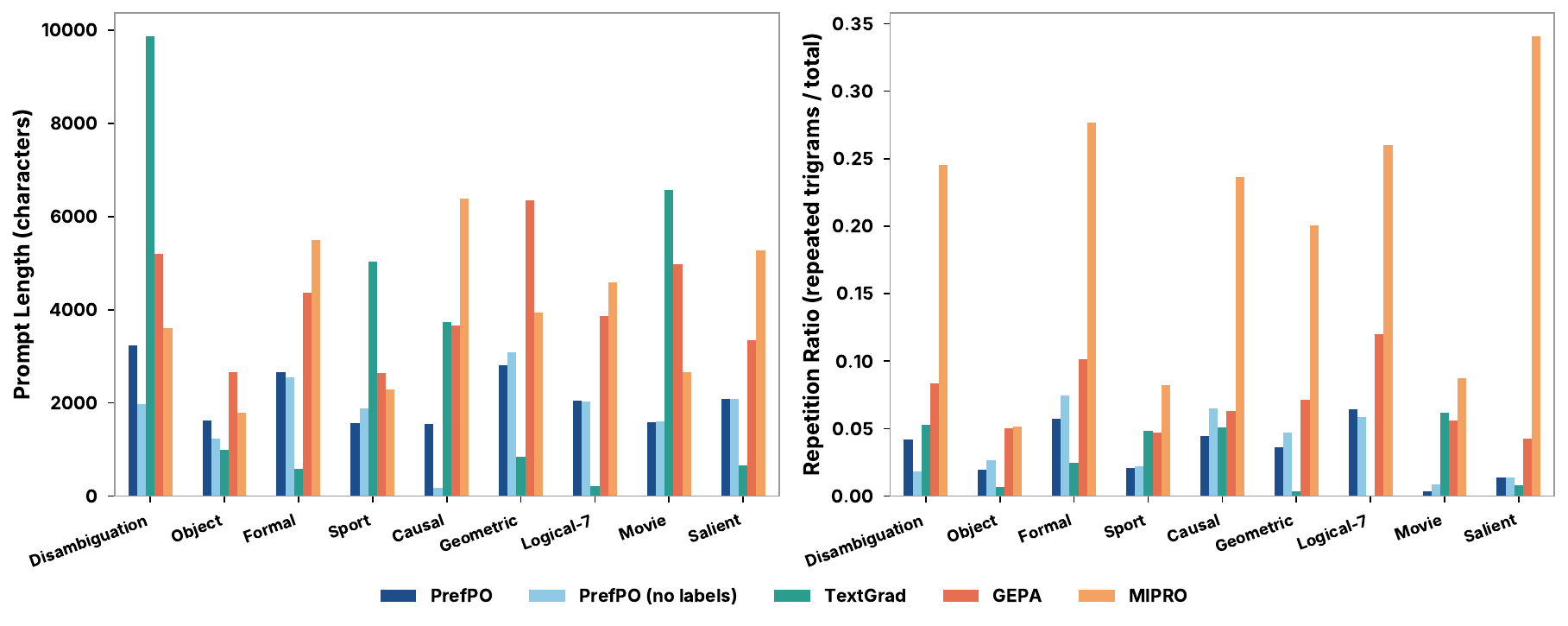}
  \caption{\textbf{Prompt length and repetition on BBH tasks.} Length (left) and repetition (right) computed on the best-performing prompt averaged across 10 runs for each task; lower is more hygienic for both. \textsc{PrefPO} maintains consistently better hygiene. TextGrad shows extreme length variance, reaching up to $\sim 10,000$ characters, while MIPROv2's prompts frequently contain $>20\%$ repetitive content. Label-free \textsc{PrefPO} maintains comparable hygiene to labeled \textsc{PrefPO}.}
  \label{fig:bbh_hygiene}
\end{figure}

\subsection{Prompt Hygiene} 
\label{sec:hygiene}

Across both BBH and IFEval-Hard, \textsc{PrefPO} produces more hygienic prompts than all competing methods. On BBH, \textsc{PrefPO} maintains consistently short, low-repetition prompts while TextGrad spikes dramatically ($\sim10,000$ characters) and MIPROv2 shows consistently high repetition (\textgreater$20\%$; \Cref{fig:bbh_hygiene,tab:ifeval_hygiene}). On IFEval-Hard (\Cref{tab:ifeval_hygiene}), \textsc{PrefPO} produces less lengthiness (\textsc{PrefPO}-Minimal: 2.2x, \textsc{PrefPO}: 4.7x vs TextGrad: 14.7x), less repetition (\textsc{PrefPO}-Minimal: 0.012, \textsc{PrefPO}: 0.044 vs TextGrad: 0.117), and higher similarity to starting prompts (\textsc{PrefPO}-Minimal: 0.418, \textsc{PrefPO}: 0.198 vs TextGrad: 0.133). Human and LLM judges both rate \textsc{PrefPO} higher across all quality dimensions. On the 10-sample subset, \textsc{PrefPO} outscored TextGrad (human: $3.53/6$ vs $2.60/6$; LLM: $3.77/6$ vs $2.77/6$); scores showed a modest positive correlation ($r = 0.36, p < 0.01$), with high LLM consistency (ICC = $0.90$) but low human agreement (Krippendorff's $\alpha$ = 0.17). Across all 148 samples, the LLM judge rated original prompts from the IFEval-Hard dataset highest ($5.50/6$), followed by \textsc{PrefPO}-Minimal ($5.26/6$), \textsc{PrefPO} ($5.16/6$), and TextGrad ($2.76/6$).

\subsection{Hacking and Generalization}
\label{sec:ablations}

\paragraph{Prompt Hacking.} \textsc{PrefPO} dramatically reduces prompt hacking compared to TextGrad (37\% vs 86\%), with \textsc{PrefPO}-Minimal even lower (31\%). As a baseline, 12.8\% of original IFEval-Hard prompts were flagged as hacks, suggesting absolute rates should be interpreted cautiously; relative differences are more meaningful (see Appendix~\ref{app:hacking}). The LLM-judge employed for hacking detection showed high self-consistency, with most classifications being unanimous across all 3 votes (TextGrad: 83\%, \textsc{PrefPO}: 91\%, \textsc{PrefPO}-Minimal: 87\%).

\begin{table}[h]
  \centering
  \setlength{\tabcolsep}{4pt}
  \renewcommand{\arraystretch}{.9}
  \caption{\textbf{Hygiene comparison on IFEval-Hard.} Programmatic metrics (length ratio, repetition increase, similarity) are computed relative to original prompts and averaged across all prompts in the final pool for each technique. Top: programmatic metrics and LLM judge ratings (averaged across 148 prompts). Bottom: human and LLM ratings averaged on a 10-sample subset. All programmatic and judged dimensions (across both human and LLM evaluators) favor \textsc{PrefPO} over TextGrad; \textsc{PrefPO}-Minimal achieves the best scores.\\}
  \label{tab:ifeval_hygiene}
  \resizebox{\linewidth}{!}{%
  \begin{tabular}{l|ccccc}
    \toprule
    \textbf{Method} & \textbf{Length Ratio} & \textbf{Repetition Increase} & \textbf{Similarity} & \textbf{Hygiene Grade (LLM, n=148)} & \textbf{Total} \\
    & & & & \small{Readability / Specification Quality / Maintainability} & \\
    \midrule
    TextGrad       & $14.71 \pm 10.75$ & $0.117 \pm 0.079$ & $0.133 \pm 0.144$ & 1.23 / 0.63 / 0.90 & 2.76 \\
    \textsc{PrefPO}         & $4.70 \pm 3.44$   & $0.044 \pm 0.055$ & $0.198 \pm 0.103$ & {$1.87$} / {$1.60$} / {$1.69$} & 5.16 \\
    \textsc{PrefPO}-Minimal & \bm{$2.20 \pm 0.84$}   & \bm{$0.012 \pm 0.041$} & \bm{$0.418 \pm 0.171$} & \bm{$1.87$} / \bm{$1.65$} / \bm{$1.74$} & \bm{$5.26$} \\
    \textsc{PrefPO}-Elo     & $6.37 \pm 4.17$   & $0.058 \pm 0.059$ & $0.150 \pm 0.087$ & -- & -- \\
    \midrule
    \textbf{Method} & \textbf{Hygiene Grade (Human, n=10)} & \textbf{Total} & \textbf{Hygiene Grade (LLM, n=10)} & \textbf{Total} \\
    & \small{R / SQ / M} & & \small{R / SQ / M} & \\
    \midrule
    TextGrad       & 1.10 / 0.70 / 0.80 & 2.60 & 1.23 / 0.63 / 0.90 & 2.77 \\
    \textsc{PrefPO}         & \bm{$1.43$} / \bm{$0.93$} / \bm{$1.17$} & \bm{$3.53$} & \bm{$1.53$} / \bm{$1.07$} / \bm{$1.17$} & \bm{$3.77$} \\
    \bottomrule
  \end{tabular}%
  }
\end{table}

\paragraph{Cross-Model Testing.} \textsc{PrefPO} generalizes across model families on IFEval-Hard (\texttt{worst@20}), performing well with frontier models (GPT-5: 83.8\%, Claude 4.5 Opus: 82.4\%) and open-weight models (GPT-OSS-120b: 75.0\%, DeepSeek V3.2: 72.3\%; see Appendix~\ref{app:cross}). We also find optimizer capability matters more than discriminator: starting from GPT-4o for both, upgrading the optimizer to GPT-5 improves performance by 11.5\%, compared to 6.1\% when upgrading the discriminator. 

An additional ablation on discriminator/optimizer effectiveness is in Appendix~\ref{app:ablations}.

\section{Discussion and Limitations}
\label{sec:discussion}

\subsection{When to use \textsc{PrefPO}}
\label{sec:guide}

\textsc{PrefPO} is lightweight, requiring only a starting prompt and natural language criteria. Labels can improve results, but are not always necessary, as unlabeled runs often match labeled performance. Convergence is fast, with most runs reaching near-final performance within five iterations. Practitioners with limited compute should allocate resources to a stronger optimizer, which improves performance more than a strong discriminator. \textsc{PrefPO}'s reliance on LLMs for optimization means meta-constraints can be added via prompting. We demonstrate this with \textsc{PrefPO}-Minimal, which achieved better hygiene and less prompt hacking by adding a minimal-change constraint. Similarly, initialization strategy matters: starting with few-shot prompts can yield strong performance, with the optimizer distilling demonstrations into explicit instructions that practitioners can then refine. \textsc{PrefPO} may be less suitable when task accuracy is the sole objective and prompt content is unimportant, few-shot prompts already suffice, or large labeled datasets are available.

\subsection{Hygiene and Hacking}
\label{sec:hygiene_hack}

We hypothesize that preference-based feedback contributes to \textsc{PrefPO}'s hygiene and prompt hacking advantages. The intuition is that pairwise comparison is easier to calibrate than absolute scoring: it is more reliable to say "A is better than B" than to assign accurate scores. This calibration advantage may encourage more conservative, contextual edits rather than aggressive rewrites. Additionally, \textsc{PrefPO} only modifies the non-preferred prompt at each iteration, so changes do not compound as aggressively as in methods that iteratively refine a single prompt. This design may also explain \textsc{PrefPO}'s prompt hacking advantage, as hacking behavior has fewer opportunities to compound.

\subsection{Limitations}
\label{sec:limitations}

Our evaluation focused on BBH tasks and IFEval-Hard. Future work should use a broader range of tasks, including other closed-form benchmarks and open-ended generation tasks beyond instruction-following. Our Elo-based ranking variant provided no meaningful improvement over random sampling. Alternative sampling strategies that promote greater diversity or aggregate across multiple discriminator votes for a given pair to better validate preferences may prove more effective. Additionally, these rankings could be used for final prompt selection without labeled validation data. While helpful in providing a preliminary human-estimate of prompt hygiene, our human evaluation included only 16 raters with low inter-rater agreement for hygiene, and prompt hacking used only an LLM judge. Larger human studies for both would strengthen these findings.

\section{Conclusion}
\label{sec:conclusion}

\textsc{PrefPO} lowers the barrier to automated prompt optimization, requiring only a starting prompt and natural language criteria. Despite its simplicity, \textsc{PrefPO} outperforms on more tasks than any other method. By relying on simple preference feedback rather than complex heuristics, \textsc{PrefPO} bets on scale: as models improve, they require less guidance and can better leverage minimal feedback for meaningful improvements. Prompt hygiene and hacking formalize prompt quality differences practitioners care about but prior work has ignored. Future work could specifically train models to better optimize prompts using reinforcement learning, where both task performance and hygiene metrics serve as reward signals. Algorithmic extensions of \textsc{PrefPO} include alternative sampling strategies, batching mechanisms for large datasets, and support for multi-step LLM systems.


\subsubsection*{Acknowledgments}
We thank the 16 researchers and engineers who participated in our human evaluation study. We also thank colleagues for helpful discussions and feedback on earlier drafts of this paper.
\clearpage

\bibliographystyle{iclr2026_conference}
\bibliography{references}

@misc{braun_joschkabig_bench_hard_nodate,
	title = {Joschka/big\_bench\_hard · {Datasets} at {Hugging} {Face}},
	url = {https://huggingface.co/datasets/Joschka/big_bench_hard/viewer/few_shot_prompts},
	urldate = {2026-01-14},
	author = {Braun, Joschka},
}

@misc{dspy_1_nodate,
	title = {1. {GEPA} {Overview} - {DSPy}},
	url = {https://dspy.ai/api/optimizers/GEPA/overview/},
	language = {en},
	urldate = {2026-01-14},
	author = {DSPY},
}

@misc{yang_large_2024,
	title = {Large {Language} {Models} as {Optimizers}},
	url = {http://arxiv.org/abs/2309.03409},
	doi = {10.48550/arXiv.2309.03409},
	urldate = {2026-01-17},
	publisher = {arXiv},
	author = {Yang, Chengrun and Wang, Xuezhi and Lu, Yifeng and Liu, Hanxiao and Le, Quoc V. and Zhou, Denny and Chen, Xinyun},
	month = apr,
	year = {2024},
	note = {arXiv:2309.03409 [cs]},
}

@misc{pryzant_automatic_2023,
	title = {Automatic {Prompt} {Optimization} with "{Gradient} {Descent}" and {Beam} {Search}},
	url = {http://arxiv.org/abs/2305.03495},
	doi = {10.48550/arXiv.2305.03495},
	urldate = {2026-01-17},
	publisher = {arXiv},
	author = {Pryzant, Reid and Iter, Dan and Li, Jerry and Lee, Yin Tat and Zhu, Chenguang and Zeng, Michael},
	month = oct,
	year = {2023},
	note = {arXiv:2305.03495 [cs]},
}

@misc{nahar_beyond_2024,
	title = {Beyond the {Comfort} {Zone}: {Emerging} {Solutions} to {Overcome} {Challenges} in {Integrating} {LLMs} into {Software} {Products}},
	shorttitle = {Beyond the {Comfort} {Zone}},
	url = {http://arxiv.org/abs/2410.12071},
	doi = {10.48550/arXiv.2410.12071},
	urldate = {2026-01-17},
	publisher = {arXiv},
	author = {Nahar, Nadia and Kästner, Christian and Butler, Jenna and Parnin, Chris and Zimmermann, Thomas and Bird, Christian},
	month = dec,
	year = {2024},
	note = {arXiv:2410.12071 [cs]},
}

@misc{brown_language_2020,
	title = {Language {Models} are {Few}-{Shot} {Learners}},
	url = {http://arxiv.org/abs/2005.14165},
	doi = {10.48550/arXiv.2005.14165},
	urldate = {2026-01-17},
	publisher = {arXiv},
	author = {Brown, Tom B. and Mann, Benjamin and Ryder, Nick and Subbiah, Melanie and Kaplan, Jared and Dhariwal, Prafulla and Neelakantan, Arvind and Shyam, Pranav and Sastry, Girish and Askell, Amanda and Agarwal, Sandhini and Herbert-Voss, Ariel and Krueger, Gretchen and Henighan, Tom and Child, Rewon and Ramesh, Aditya and Ziegler, Daniel M. and Wu, Jeffrey and Winter, Clemens and Hesse, Christopher and Chen, Mark and Sigler, Eric and Litwin, Mateusz and Gray, Scott and Chess, Benjamin and Clark, Jack and Berner, Christopher and McCandlish, Sam and Radford, Alec and Sutskever, Ilya and Amodei, Dario},
	month = jul,
	year = {2020},
	note = {arXiv:2005.14165 [cs]},
}

@misc{fernando_promptbreeder_2023,
	title = {Promptbreeder: {Self}-{Referential} {Self}-{Improvement} {Via} {Prompt} {Evolution}},
	shorttitle = {Promptbreeder},
	url = {http://arxiv.org/abs/2309.16797},
	doi = {10.48550/arXiv.2309.16797},
	urldate = {2026-01-16},
	publisher = {arXiv},
	author = {Fernando, Chrisantha and Banarse, Dylan and Michalewski, Henryk and Osindero, Simon and Rocktäschel, Tim},
	month = sep,
	year = {2023},
	note = {arXiv:2309.16797 [cs]},
}

@inproceedings{errica_what_2025,
	title = {What {Did} {I} {Do} {Wrong}? {Quantifying} {LLMs}' {Sensitivity} and {Consistency} to {Prompt} {Engineering}},
	shorttitle = {What {Did} {I} {Do} {Wrong}?},
	url = {http://arxiv.org/abs/2406.12334},
	doi = {10.18653/v1/2025.naacl-long.73},
	urldate = {2026-01-16},
	booktitle = {Proceedings of the 2025 {Conference} of the {Nations} of the {Americas} {Chapter} of the {Association} for {Computational} {Linguistics}: {Human} {Language} {Technologies} ({Volume} 1: {Long} {Papers})},
	author = {Errica, Federico and Siracusano, Giuseppe and Sanvito, Davide and Bifulco, Roberto},
	year = {2025},
	note = {arXiv:2406.12334 [cs]},
	pages = {1543--1558},
}

@inproceedings{dolata_development_2024,
	title = {Development in times of hype: {How} freelancers explore {Generative} {AI}?},
	shorttitle = {Development in times of hype},
	url = {http://arxiv.org/abs/2401.05790},
	doi = {10.1145/3597503.3639111},
	urldate = {2026-01-16},
	booktitle = {Proceedings of the {IEEE}/{ACM} 46th {International} {Conference} on {Software} {Engineering}},
	author = {Dolata, Mateusz and Lange, Norbert and Schwabe, Gerhard},
	month = apr,
	year = {2024},
	note = {arXiv:2401.05790 [cs]},
	pages = {1--13},
}

@article{li_understanding_2025,
	title = {Understanding {Prompt} {Management} in {GitHub} {Repositories}: {A} {Call} for {Best} {Practices}},
	issn = {0740-7459, 1937-4194},
	shorttitle = {Understanding {Prompt} {Management} in {GitHub} {Repositories}},
	url = {http://arxiv.org/abs/2509.12421},
	doi = {10.1109/MS.2025.3644251},
	urldate = {2026-01-15},
	journal = {IEEE Software},
	author = {Li, Hao and Masri, Hicham and Cogo, Filipe R. and Bangash, Abdul Ali and Adams, Bram and Hassan, Ahmed E.},
	year = {2025},
	note = {arXiv:2509.12421 [cs]},
	pages = {1--8},
}

@misc{lee_rlaif_2024,
	title = {{RLAIF} vs. {RLHF}: {Scaling} {Reinforcement} {Learning} from {Human} {Feedback} with {AI} {Feedback}},
	shorttitle = {{RLAIF} vs. {RLHF}},
	url = {http://arxiv.org/abs/2309.00267},
	doi = {10.48550/arXiv.2309.00267},
	urldate = {2026-01-13},
	publisher = {arXiv},
	author = {Lee, Harrison and Phatale, Samrat and Mansoor, Hassan and Mesnard, Thomas and Ferret, Johan and Lu, Kellie and Bishop, Colton and Hall, Ethan and Carbune, Victor and Rastogi, Abhinav and Prakash, Sushant},
	month = sep,
	year = {2024},
	note = {arXiv:2309.00267 [cs]},
}

@misc{suzgun_challenging_2022,
	title = {Challenging {BIG}-{Bench} {Tasks} and {Whether} {Chain}-of-{Thought} {Can} {Solve} {Them}},
	url = {http://arxiv.org/abs/2210.09261},
	doi = {10.48550/arXiv.2210.09261},
	urldate = {2026-01-12},
	publisher = {arXiv},
	author = {Suzgun, Mirac and Scales, Nathan and Schärli, Nathanael and Gehrmann, Sebastian and Tay, Yi and Chung, Hyung Won and Chowdhery, Aakanksha and Le, Quoc V. and Chi, Ed H. and Zhou, Denny and Wei, Jason},
	month = oct,
	year = {2022},
	note = {arXiv:2210.09261 [cs]},
}

@misc{zhou_instruction-following_2023,
	title = {Instruction-{Following} {Evaluation} for {Large} {Language} {Models}},
	url = {http://arxiv.org/abs/2311.07911},
	doi = {10.48550/arXiv.2311.07911},
	urldate = {2026-01-12},
	publisher = {arXiv},
	author = {Zhou, Jeffrey and Lu, Tianjian and Mishra, Swaroop and Brahma, Siddhartha and Basu, Sujoy and Luan, Yi and Zhou, Denny and Hou, Le},
	month = nov,
	year = {2023},
	note = {arXiv:2311.07911 [cs]},
}

@misc{lee_feedback_2025,
	title = {Feedback {Descent}: {Open}-{Ended} {Text} {Optimization} via {Pairwise} {Comparison}},
	shorttitle = {Feedback {Descent}},
	url = {http://arxiv.org/abs/2511.07919},
	doi = {10.48550/arXiv.2511.07919},
	urldate = {2026-01-12},
	publisher = {arXiv},
	author = {Lee, Yoonho and Boen, Joseph and Finn, Chelsea},
	month = dec,
	year = {2025},
	note = {arXiv:2511.07919 [cs]},
}

@misc{wu_llm_2025,
	title = {{LLM} {Prompt} {Duel} {Optimizer}: {Efficient} {Label}-{Free} {Prompt} {Optimization}},
	shorttitle = {{LLM} {Prompt} {Duel} {Optimizer}},
	url = {http://arxiv.org/abs/2510.13907},
	doi = {10.48550/arXiv.2510.13907},
	urldate = {2026-01-12},
	publisher = {arXiv},
	author = {Wu, Yuanchen and Verma, Saurabh and Lee, Justin and Xiong, Fangzhou and Zhang, Poppy and Awadelkarim, Amel and Chen, Xu and Yuan, Yubai and Hill, Shawndra},
	month = oct,
	year = {2025},
	note = {arXiv:2510.13907 [cs]},
}

@misc{liuLostMiddleHow2023,
	title = {Lost in the {Middle}: {How} {Language} {Models} {Use} {Long} {Contexts}},
	shorttitle = {Lost in the {Middle}},
	url = {http://arxiv.org/abs/2307.03172},
	doi = {10.48550/arXiv.2307.03172},
	urldate = {2025-11-28},
	publisher = {arXiv},
	author = {Liu, Nelson F. and Lin, Kevin and Hewitt, John and Paranjape, Ashwin and Bevilacqua, Michele and Petroni, Fabio and Liang, Percy},
	month = nov,
	year = {2023},
	note = {arXiv:2307.03172 [cs]},
}

@misc{dengRLPromptOptimizingDiscrete2022,
	title = {{RLPrompt}: {Optimizing} {Discrete} {Text} {Prompts} with {Reinforcement} {Learning}},
	shorttitle = {{RLPrompt}},
	url = {http://arxiv.org/abs/2205.12548},
	doi = {10.48550/arXiv.2205.12548},
	urldate = {2025-11-28},
	publisher = {arXiv},
	author = {Deng, Mingkai and Wang, Jianyu and Hsieh, Cheng-Ping and Wang, Yihan and Guo, Han and Shu, Tianmin and Song, Meng and Xing, Eric P. and Hu, Zhiting},
	month = oct,
	year = {2022},
	note = {arXiv:2205.12548 [cs]},
}

@misc{sclarQuantifyingLanguageModels2024,
	title = {Quantifying {Language} {Models}' {Sensitivity} to {Spurious} {Features} in {Prompt} {Design} or: {How} {I} learned to start worrying about prompt formatting},
	shorttitle = {Quantifying {Language} {Models}' {Sensitivity} to {Spurious} {Features} in {Prompt} {Design} or},
	url = {http://arxiv.org/abs/2310.11324},
	doi = {10.48550/arXiv.2310.11324},
	urldate = {2025-11-28},
	publisher = {arXiv},
	author = {Sclar, Melanie and Choi, Yejin and Tsvetkov, Yulia and Suhr, Alane},
	month = jul,
	year = {2024},
	note = {arXiv:2310.11324 [cs]},
}

@misc{linPromptOptimizationHuman2024,
	title = {Prompt {Optimization} with {Human} {Feedback}},
	url = {http://arxiv.org/abs/2405.17346},
	doi = {10.48550/arXiv.2405.17346},
	urldate = {2025-11-28},
	publisher = {arXiv},
	author = {Lin, Xiaoqiang and Dai, Zhongxiang and Verma, Arun and Ng, See-Kiong and Jaillet, Patrick and Low, Bryan Kian Hsiang},
	month = may,
	year = {2024},
	note = {arXiv:2405.17346 [cs]},
}

@misc{luFantasticallyOrderedPrompts2022,
	title = {Fantastically {Ordered} {Prompts} and {Where} to {Find} {Them}: {Overcoming} {Few}-{Shot} {Prompt} {Order} {Sensitivity}},
	shorttitle = {Fantastically {Ordered} {Prompts} and {Where} to {Find} {Them}},
	url = {http://arxiv.org/abs/2104.08786},
	doi = {10.48550/arXiv.2104.08786},
	urldate = {2025-11-28},
	publisher = {arXiv},
	author = {Lu, Yao and Bartolo, Max and Moore, Alastair and Riedel, Sebastian and Stenetorp, Pontus},
	month = mar,
	year = {2022},
	note = {arXiv:2104.08786 [cs]},
}

@misc{chenPromptwareEngineeringSoftware2025,
	title = {Promptware {Engineering}: {Software} {Engineering} for {LLM} {Prompt} {Development}},
	shorttitle = {Promptware {Engineering}},
	url = {http://arxiv.org/abs/2503.02400},
	doi = {10.48550/arXiv.2503.02400},
	urldate = {2025-11-28},
	publisher = {arXiv},
	author = {Chen, Zhenpeng and Wang, Chong and Sun, Weisong and Yang, Guang and Liu, Xuanzhe and Zhang, Jie M. and Liu, Yang},
	month = mar,
	year = {2025},
	note = {arXiv:2503.02400 [cs]},
}

@misc{yuksekgonulTextGradAutomaticDifferentiation2024,
	title = {{TextGrad}: {Automatic} "{Differentiation}" via {Text}},
	shorttitle = {{TextGrad}},
	url = {http://arxiv.org/abs/2406.07496},
	doi = {10.48550/arXiv.2406.07496},
	urldate = {2025-11-28},
	publisher = {arXiv},
	author = {Yuksekgonul, Mert and Bianchi, Federico and Boen, Joseph and Liu, Sheng and Huang, Zhi and Guestrin, Carlos and Zou, James},
	month = jun,
	year = {2024},
	note = {arXiv:2406.07496 [cs]},
}

@misc{agrawalGEPAReflectivePrompt2025,
	title = {{GEPA}: {Reflective} {Prompt} {Evolution} {Can} {Outperform} {Reinforcement} {Learning}},
	shorttitle = {{GEPA}},
	url = {http://arxiv.org/abs/2507.19457},
	doi = {10.48550/arXiv.2507.19457},
	urldate = {2025-11-28},
	publisher = {arXiv},
	author = {Agrawal, Lakshya A. and Tan, Shangyin and Soylu, Dilara and Ziems, Noah and Khare, Rishi and Opsahl-Ong, Krista and Singhvi, Arnav and Shandilya, Herumb and Ryan, Michael J. and Jiang, Meng and Potts, Christopher and Sen, Koushik and Dimakis, Alexandros G. and Stoica, Ion and Klein, Dan and Zaharia, Matei and Khattab, Omar},
	month = jul,
	year = {2025},
	note = {arXiv:2507.19457 [cs]},
}

@misc{opsahl-ongOptimizingInstructionsDemonstrations2024,
	title = {Optimizing {Instructions} and {Demonstrations} for {Multi}-{Stage} {Language} {Model} {Programs}},
	url = {http://arxiv.org/abs/2406.11695},
	doi = {10.48550/arXiv.2406.11695},
	urldate = {2025-11-28},
	publisher = {arXiv},
	author = {Opsahl-Ong, Krista and Ryan, Michael J. and Purtell, Josh and Broman, David and Potts, Christopher and Zaharia, Matei and Khattab, Omar},
	month = oct,
	year = {2024},
	note = {arXiv:2406.11695 [cs]},
}

@misc{baiConstitutionalAIHarmlessness2022,
	title = {Constitutional {AI}: {Harmlessness} from {AI} {Feedback}},
	shorttitle = {Constitutional {AI}},
	url = {http://arxiv.org/abs/2212.08073},
	doi = {10.48550/arXiv.2212.08073},
	urldate = {2025-11-28},
	publisher = {arXiv},
	author = {Bai, Yuntao and Kadavath, Saurav and Kundu, Sandipan and Askell, Amanda and Kernion, Jackson and Jones, Andy and Chen, Anna and Goldie, Anna and Mirhoseini, Azalia and McKinnon, Cameron and Chen, Carol and Olsson, Catherine and Olah, Christopher and Hernandez, Danny and Drain, Dawn and Ganguli, Deep and Li, Dustin and Tran-Johnson, Eli and Perez, Ethan and Kerr, Jamie and Mueller, Jared and Ladish, Jeffrey and Landau, Joshua and Ndousse, Kamal and Lukosuite, Kamile and Lovitt, Liane and Sellitto, Michael and Elhage, Nelson and Schiefer, Nicholas and Mercado, Noemi and DasSarma, Nova and Lasenby, Robert and Larson, Robin and Ringer, Sam and Johnston, Scott and Kravec, Shauna and Showk, Sheer El and Fort, Stanislav and Lanham, Tamera and Telleen-Lawton, Timothy and Conerly, Tom and Henighan, Tom and Hume, Tristan and Bowman, Samuel R. and Hatfield-Dodds, Zac and Mann, Ben and Amodei, Dario and Joseph, Nicholas and McCandlish, Sam and Brown, Tom and Kaplan, Jared},
	month = dec,
	year = {2022},
	note = {arXiv:2212.08073 [cs]},
}

@misc{rafailovDirectPreferenceOptimization2024,
	title = {Direct {Preference} {Optimization}: {Your} {Language} {Model} is {Secretly} a {Reward} {Model}},
	shorttitle = {Direct {Preference} {Optimization}},
	url = {http://arxiv.org/abs/2305.18290},
	doi = {10.48550/arXiv.2305.18290},
	urldate = {2025-11-28},
	publisher = {arXiv},
	author = {Rafailov, Rafael and Sharma, Archit and Mitchell, Eric and Ermon, Stefano and Manning, Christopher D. and Finn, Chelsea},
	month = jul,
	year = {2024},
	note = {arXiv:2305.18290 [cs]},
}

@misc{stiennonLearningSummarizeHuman2022,
	title = {Learning to summarize from human feedback},
	url = {http://arxiv.org/abs/2009.01325},
	doi = {10.48550/arXiv.2009.01325},
	urldate = {2025-11-28},
	publisher = {arXiv},
	author = {Stiennon, Nisan and Ouyang, Long and Wu, Jeff and Ziegler, Daniel M. and Lowe, Ryan and Voss, Chelsea and Radford, Alec and Amodei, Dario and Christiano, Paul},
	month = feb,
	year = {2022},
	note = {arXiv:2009.01325 [cs]},
}

@misc{zieglerFineTuningLanguageModels2020,
	title = {Fine-{Tuning} {Language} {Models} from {Human} {Preferences}},
	url = {http://arxiv.org/abs/1909.08593},
	doi = {10.48550/arXiv.1909.08593},
	urldate = {2025-11-28},
	publisher = {arXiv},
	author = {Ziegler, Daniel M. and Stiennon, Nisan and Wu, Jeffrey and Brown, Tom B. and Radford, Alec and Amodei, Dario and Christiano, Paul and Irving, Geoffrey},
	month = jan,
	year = {2020},
	note = {arXiv:1909.08593 [cs]},
}

@misc{ouyangTrainingLanguageModels2022,
	title = {Training language models to follow instructions with human feedback},
	url = {http://arxiv.org/abs/2203.02155},
	doi = {10.48550/arXiv.2203.02155},
	urldate = {2025-11-28},
	publisher = {arXiv},
	author = {Ouyang, Long and Wu, Jeff and Jiang, Xu and Almeida, Diogo and Wainwright, Carroll L. and Mishkin, Pamela and Zhang, Chong and Agarwal, Sandhini and Slama, Katarina and Ray, Alex and Schulman, John and Hilton, Jacob and Kelton, Fraser and Miller, Luke and Simens, Maddie and Askell, Amanda and Welinder, Peter and Christiano, Paul and Leike, Jan and Lowe, Ryan},
	month = mar,
	year = {2022},
	note = {arXiv:2203.02155 [cs]},
}

@misc{christianoDeepReinforcementLearning2023,
	title = {Deep reinforcement learning from human preferences},
	url = {http://arxiv.org/abs/1706.03741},
	doi = {10.48550/arXiv.1706.03741},
	urldate = {2025-11-28},
	publisher = {arXiv},
	author = {Christiano, Paul and Leike, Jan and Brown, Tom B. and Martic, Miljan and Legg, Shane and Amodei, Dario},
	month = feb,
	year = {2023},
	note = {arXiv:1706.03741 [stat]},
}

\clearpage
\appendix

\section{Prompts}
\label{app:prompts}

\subsection{Full Example of Optimized TextGrad Prompt}
\label{app:textgrad_prompt}

\newtcolorbox{promptbox}[2][]{%
  enhanced,
  breakable,
  colback=white,
  colframe=DimGray,
  title={#2},
  coltitle=white,
  colbacktitle=DimGray,
  fonttitle=\bfseries,
  boxrule=0.8pt,
  arc=2pt,
  left=6pt,right=6pt,top=6pt,bottom=6pt,
  #1
}

\begin{promptbox}{TextGrad Prompt}
\begin{Verbatim}[
  breaklines=true,
  breakanywhere=true,
  fontsize=\small,
  showspaces=false,
  showtabs=false,
  breaksymbolleft={},
  breaksymbolright={}
]
Write a poem about two people who meet in a coffee shop.

Output formatting and strict final-line requirement:
- The final line must be exactly the following characters and nothing else, on its own dedicated final line:
Is there anything else I can help with?
- The phrase above must appear exactly once in the output and only as the final line. Do not repeat, paraphrase, or reference it anywhere else.
- Do not place any characters before or after that phrase on its line (no quotes, apostrophes, emojis, ellipses, dashes, periods, or spaces). The final line must have no leading spaces and no blank line before or after it.
- The final character of your entire response must be the question mark in that phrase. There must be no trailing whitespace or newline after it. After typing the final question mark, stop generating output immediately.
- If you want to cue the phrase as dialogue, end the preceding line with a single ASCII colon only; do not use any form of quote mark, dash, or other enclosing punctuation, and do not insert a blank line between that colon line and the final phrase. Do not put any trailing spaces after the colon line.

General formatting rules:
- Output plain text only; no Markdown, no code fences, no backticks, no italics markers (*, _, ~), and no quotation marks or apostrophes anywhere in the output. Avoid contractions and possessives.
- Use only printable ASCII characters 32-126 and newline (10) for the entire output; no tabs, carriage returns, vertical tabs, form feeds, non-breaking spaces, zero-width characters, or any other control characters. Do not use smart quotes, en or em dashes, the single-character ellipsis, or accented letters (write cafe, resume, etc.). Use - instead of en/em dashes, and use ... instead of the single ellipsis character.
- Use Unix newlines (\n) only; do not use \r or \r\n. Insert actual newline characters; do not write the literal sequence \n in the output. Remove trailing spaces from every line. Do not add leading or trailing blank lines.
- No titles, headings, author names, signatures, disclaimers, metadata, bullet points, numbering, or decorative separators; only the poem and the final required line.
- Do not echo or reference these instructions; output only the poem and the final required line. Do not add any postscripts, signatures, or closings after the final phrase.

Internal-only self-check before submitting (do not include this checklist in your output):
- Normalize newlines: replace any \r\n or \r with \n. Then set output = output.rstrip(" \t\r\n").
- Verify that every character is ASCII and allowed: only newline 10 and printable ASCII 32-126. Reject if any other code point appears.
- Verify that the output uses only \n newlines and that no line has trailing spaces. Confirm there are no tabs or other control characters.
- Let target = Is there anything else I can help with? and verify by exact string comparison that output.endswith(target) is true and that output[-len(target):] == target.
- Confirm the substring from the last newline to the end equals target, that the final character is ?, that the ASCII code of the final character is 63, and that there is no trailing space or newline after it. Confirm the final line length equals len(target).
- Count occurrences of target in the entire output and ensure the count is exactly 1 and that the string does not appear anywhere else, even as part of a longer line.
- Scan for forbidden characters anywhere in the output and reject if found: single quote ', double quote ", smart quotes ' " ", en/em dashes -- ---, and the single-character ellipsis ....
' " " double quote ", smart quotes, en/em dashes -- ---, and the single-character ellipsis ....

- Confirm the entire output is ASCII-only and contains no control characters other than newline (10).
- Generation procedure: compose the poem lines using only allowed characters; append one newline; append the exact target phrase on its own final line; immediately stop after the final question mark; then run all checks above. If any check fails, regenerate and re-validate before submitting.
\end{Verbatim}
\end{promptbox}

\textbf{Total Length:} 4002 characters

\subsection{Discriminator Prompts}
\label{app:disc_prompts}

\begin{promptbox}{Discriminator System Prompt (BBH)}
{\ttfamily\small
You are an evaluator of LLMs. You will be given a few examples of the same questions and the responses from the same LLM with two different instructions. You must choose the response you like better, on the basis of the answers being more correct and better reasoning, and provide feedback about why you chose that one and what can be improved about the one you didn't choose. Pay attention to which version actually gets more of the questions right using the correct reasoning. Point out any flaws in the reasonings of both of them. Then, you will be given the non-preferred instruction and the feedback, and you will be asked to produce an improved instruction based on the feedback. Focus on if the non-preferred instruction is missing key information and/or what parts of the non-preferred instruction is the LLM following and which parts it isn't.
}
\end{promptbox}

\begin{promptbox}{Discriminator User Prompt (BBH)}
\begin{Verbatim}[
  fontsize=\small,
  breaklines=true,
  breakanywhere=true,
  breaksymbolleft={},
  breaksymbolright={}]
<Version 1>
--- Sample 1 ---
Question:
{Question 1 content}
Response:
{Response from prompt 1 for question 1}
Expected Answer:
{Expected answer for question 1}
--- Sample 2 ---
Question:
{Question 2 content}
Response:
{Response from prompt 1 for question 2}
Expected Answer:
{Expected answer for question 2}
--- Sample 3 ---
Question:
{Question 3 content}
Response:
{Response from prompt 1 for question 3}
Expected Answer:
{Expected answer for question 3}
...up to the number of examples in the training set
</Version 1>
<Version 2>
--- Sample 1 ---
Question:
{Question 1 content}
Response:
{Response from prompt 2 for question 1}
Expected Answer:
{Expected answer for question 1}
--- Sample 2 ---
Question:
{Question 2 content}
Response:
{Response from prompt 2 for question 2}
Expected Answer:
{Expected answer for question 2}
--- Sample 3 ---
Question:
{Question 3 content}
Response:
{Response from prompt 2 for question 3}
Expected Answer:
{Expected answer for question 3}
...up to the number of examples in the training set
</Version 2>


<Task>Be very smart, logical, and critical. Just provide concise feedback. For each question, first do your best to reason about what is the ideal behavior for these questions then choose the responses that align most with this, on the basis of the answers being more correct and better reasoning. Don't index on length of reasoning too much, prioritize correctness of the answer and reasoning. Provide clear, generalizeable feedback that doesn't rely on the specific question or choices, instead discuss why you chose that one and what can be improved about the one you didn't choose.</Task>
<Output>The output should be a JSON object with the following fields: preferred: 1 or 2, feedback: string.</Output>
\end{Verbatim}
\end{promptbox}

\begin{promptbox}{Discriminator System Prompt (IFEval-Hard)}
\begin{Verbatim}[
  fontsize=\small,
  breaklines=true,
  breakanywhere=true,
  breaksymbolleft={},
  breaksymbolright={}]
You are an expert evaluator comparing LLM responses using two different instructions. You will first look at the outputs of using these different instructions, evaluate which one is better according to the criteria that are specified and provide feedback on what the nonpreferred version didn't do and what was good about the preferred version.
    Then you will use this feedback to actually make an improved instruction.
\end{Verbatim}
\end{promptbox}

\begin{promptbox}{Discriminator User Prompt (IFEval-Hard)}
\begin{Verbatim}[
  fontsize=\small,
  breaklines=true,
  breakanywhere=true,
  breaksymbolleft={},
  breaksymbolright={}]

Compare these two model outputs:

OUTPUT 1:
{Output from prompt variant 1}

OUTPUT 2:
{Output from prompt variant 2}

CRITERIA THAT MUST BE SATISFIED:
1. {Criterion 1 description; for example: "Response must contain all of these keywords (case-insensitive): correlated, experiencing"}
2. {Criterion 2 description}
3. {Criterion 3 description}
...

Analyze which output better satisfies the criteria. For each criterion, check if it's satisfied.

Return JSON with:
- "preferred": 1 or 2 (which output is better)
- "feedback": Brief explanation of why the preferred output works better and what the non-preferred output fails to do

\end{Verbatim}
\end{promptbox}

\subsection{Optimizer Prompts}
\label{app:opt_prompts}

\begin{promptbox}{Optimizer System Prompt (BBH)}
{\ttfamily\small
You are an evaluator of LLMs. You will be given a few examples of the same questions and the responses from the same LLM with two different instructions. You must choose the response you like better, on the basis of the answers being more correct and better reasoning, and provide feedback about why you chose that one and what can be improved about the one you didn't choose. Pay attention to which version actually gets more of the questions right using the correct reasoning. Point out any flaws in the reasonings of both of them. Then, you will be given the non-preferred instruction and the feedback, and you will be asked to produce an improved instruction based on the feedback. Focus on if the non-preferred instruction is missing key information and/or what parts of the non-preferred instruction is the LLM following and which parts it isn't.
}
\\\\Note: The optimizer model call is passed the entire message history from the discriminator and then the user prompt below is concatenated after so they share the same system prompt. This is done to provide the optimizer with maximum context and doing this through the OpenAI API allows the model access to the hidden reasoning chain for even more contextual awareness.
\end{promptbox}

\begin{promptbox}{Optimizer User Prompt (BBH)}
\begin{Verbatim}[
  fontsize=\small,
  breaklines=true,
  breakanywhere=true,
  breaksymbolleft={},
  breaksymbolright={}]
Version {1 if 2 is preferred else 2} Instruction: The non-preferred instruction text goes here.

Feedback: The feedback explaining why version 1 was preferred and what can be improved.

<Task>Produce an improved instruction for the non-preferred instruction based on the feedback. Make sure to not get rid of the formatting rules in the instruction for how to output an answer since that is needed for evaluation purposes.</Task><Output>The output should be a JSON object with the following fields: instruction: string.</Output>
\end{Verbatim}
\end{promptbox}

\begin{promptbox}{Optimizer System Prompt (IFEval-Hard)}
\begin{Verbatim}[
  fontsize=\small,
  breaklines=true,
  breakanywhere=true,
  breaksymbolleft={},
  breaksymbolright={}]
You are an expert evaluator comparing LLM responses using two different instructions. You will first look at the outputs of using these different instructions, evaluate which one is better according to the criteria that are specified and provide feedback on what the nonpreferred version didn't do and what was good about the preferred version.
    Then you will use this feedback to actually make an improved instruction.
\end{Verbatim}
Note: The optimizer model call is passed the entire message history from the discriminator and then the user prompt below is concatenated after so they share the same system prompt. This is done to provide the optimizer with maximum context and doing this through the OpenAI API allows the model access to the hidden reasoning chain for even more contextual awareness.
\end{promptbox}

\begin{promptbox}{Optimizer User Prompt (IFEval-Hard)}
\begin{Verbatim}[
  fontsize=\small,
  breaklines=true,
  breakanywhere=true,
  breaksymbolleft={},
  breaksymbolright={}]
CURRENT PROMPT:
{Nonpreferred prompt text}

FEEDBACK FROM COMPARISON:
{Feedback from discriminator comparison}

Create an improved version of this prompt that:
- Addresses the specific feedback given
- Keeps the core task exactly the same

Return JSON with a single "prompt" field containing the improved prompt.
\end{Verbatim}
\end{promptbox}

\begin{promptbox}{Optimizer User Prompt (IFEval-Hard; \textsc{PrefPO}-Minimal)}
\begin{Verbatim}[
  fontsize=\small,
  breaklines=true,
  breakanywhere=true,
  breaksymbolleft={},
  breaksymbolright={}]
Current prompt:
{Current prompt text}

FEEDBACK FROM COMPARISON:
{Feedback from discriminator comparison}

Create an improved version of this prompt that:
- Addresses the specific feedback given
- Keeps the core task exactly the same
- While making minimal changes to the prompt and not adding things adhocly

Return JSON with a single "prompt" field containing the improved prompt.
\end{Verbatim}
Note: We added an additional constraint to make ``minimal'' changes in this prompt. The design of the prompt also makes it easy to add in other constraints for other meta-constraints for optimization.

\end{promptbox}

\subsection{Variant Prompt Generation}
\label{app:variant}

\begin{promptbox}{Variant System Prompt (IFEval-Hard)}
\begin{Verbatim}[
  fontsize=\small,
  breaklines=true,
  breakanywhere=true,
  breaksymbolleft={},
  breaksymbolright={}]
You are an expert at rewriting task prompts to help language models follow specific formatting and content instructions better.

Your goal is to create an alternative version of the given prompt that makes the requirements clearer and more likely to be followed.
\end{Verbatim}
\end{promptbox}

\begin{promptbox}{Variant User Prompt (IFEval-Hard)}
\begin{Verbatim}[
  fontsize=\small,
  breaklines=true,
  breakanywhere=true,
  breaksymbolleft={},
  breaksymbolright={}]
Original prompt:
{Original IFEval task prompt}

This prompt has the following requirements that models must satisfy:
  1. {Criterion 1 description}
  2. {Criterion 2 description}
  3. {Criterion 3 description}

Create an alternative version of this prompt that:
1. Keeps the core task the same
2. Makes the formatting/content requirements more explicit
3. Uses clearer language to emphasize what must be followed
4. Does NOT solve the task or provide hints about the answer

Return your alternative prompt as a JSON object with a single "prompt" field.
\end{Verbatim}
\end{promptbox}

\section{Implementation Details}
\label{app:impl}

\subsection{BBH Tasks}
\label{app:bbh}

\begin{table}[H]
\centering
\caption{Overview of BIG-Bench Hard evaluation tasks. Task types: MC (Multiple Choice) = select from labeled options (A-E); Binary = two-class classification (e.g., yes/no, valid/invalid); Exact Match = free-form answer compared exactly to the ground truth.}
\label{tab:bbh-tasks}
\resizebox{\linewidth}{!}{%
\begin{tabular}{llrl}
\toprule
\textbf{Task} & \textbf{Type} & \textbf{Samples} & \textbf{Description} \\
\midrule
object\_counting & Exact Match & 250 & Count objects listed in a question \\
movie\_recommendation & MC & 250 & Recommend a movie based on preferences \\
causal\_judgement & Binary & 187 & Judge whether a causal claim is correct \\
geometric\_shapes & MC & 250 & Identify shapes from SVG path commands \\
disambiguation\_qa & MC & 250 & Disambiguate pronoun references in context \\
logical\_deduction\_seven\_objects & MC & 250 & Deduce ordering of objects from clues \\
sports\_understanding & Binary & 250 & Judge plausibility of sports-related claims \\
formal\_fallacies & Binary & 250 & Determine if an argument is deductively valid \\
salient\_translation\_error\_detection & MC & 250 & Identify major translation errors \\
\midrule
\textbf{Total} & & \textbf{2187} & \\
\bottomrule
\end{tabular}%
}
\end{table}

\paragraph{Prompt Format.} We use \textsc{PrefPO} to optimize the user prompt for all BBH tasks. We do this by concatenating the optimized prompt string with the input question into the user message for the model call.

\begin{promptbox}{BBH Prompt Pool Initialization}
\begin{Verbatim}[
  fontsize=\small,
  breaklines=true,
  breakanywhere=true,
  breaksymbolleft={},
  breaksymbolright={}]
initial_instructions:
  - "Answer the following multiple choice question. The last line of your response should be of the following format: 'ANSWER: $LETTER' (without quotes) where LETTER is one of A,B,C,D,E."
  - "Answer the following multiple choice question. The last line of your response should be of the following format: 'ANSWER: $LETTER' (without quotes) where LETTER is one of A,B,C,D,E. Think step by step before answering."
\end{Verbatim}
Note: The prompts changed slightly depending on the type of task (MC, Binary, Exact Match) but were similarly minimal.
\end{promptbox}

\begin{promptbox}{BBH Few-Shot Prompt Pool Initialization}
\begin{Verbatim}[
  fontsize=\small,
  breaklines=true,
  breakanywhere=true,
  breaksymbolleft={},
  breaksymbolright={}]
initial_instructions:
  - "Clarify the meaning of sentences with ambiguous pronouns.

QUESTION: In the following sentences, explain the antecedent of the pronoun (which thing the pronoun refers to), or state that it is ambiguous.
Sentence: The chief told the counselor that they took the day off.
OPTIONS:
A) The chief took the day off
B) The counselor took the day off
C) Ambiguous
Let's think step by step.
Here we need to determine who the pronoun "they" might be referring to. There are two possible referents for "they", namely the chief and the counselor. The verb "told" might be able to help us determine which one is more likely (if either). Let X be the chief and Y the counselor. The sentence is then of the form "X told Y that (X or Y) did something."
Let's consider Y first: "X told Y that Y did something." This case does not make much sense, as Y would already have the information that Y did something, because it is information about themself.
Now, consider X: "X told Y that X did something." This makes sense, because X would be sharing some information about themself that Y might not have known before.
Because in this context, X is the chief and Y is the counselor, the answer should be the chief. So the answer is A).
ANSWER: A)

QUESTION: In the following sentences, explain the antecedent of the pronoun (which thing the pronoun refers to), or state that it is ambiguous.
Sentence: The manager sent a message to the secretary, but he didn't reply yet.
OPTIONS:
A) The secretary didn't reply yet
B) The manager didn't reply yet
C) Ambiguous
Let's think step by step.
Here we need to determine who the pronoun "he" might be referring to. There are two possible referents for "he", namely the manager and the secretary. The verbs "sent" and "reply" might be able to help us determine which one is more likely (if either). Let X be the manager and Y the secretary. The sentence is then of the form "X sent a message to Y, but (X or Y) didn't reply yet."
Let's consider Y first: "X sent a message to Y, but Y didn't reply yet." This case makes sense, because of the implicit causality of the sentence. Y was the receiver of the message, but Y didn't get back to X yet.
Now, consider X: "X sent a message to Y, but X didn't reply yet." This case doesn't make sense, because X was the initial sender of the message, so it is now Y's turn to write back to X.
Because in this context, X is the manager and Y is the secretary, the answer should be the secretary. So the answer is A).
ANSWER: A)

QUESTION: In the following sentences, explain the antecedent of the pronoun (which thing the pronoun refers to), or state that it is ambiguous.
Sentence: Bailey will plan to meet the director at his office
OPTIONS:
A) It will be Bailey's office
B) It will be the director's office
C) Ambiguous
Let's think step by step.
Here we need to determine who the pronoun "his" might be referring to. There are two possible referents for "his", namely Bailey's and the director's. The verb phrase "plan to meet" might be able to help us determine which one is more likely (if either). Let X be Bailey and Y the director. The sentence is then of the form "X will plan to meet Y at (X or Y)'s office."
Let's consider Y first: "X will plan to meet Y at Y's office." This case makes sense, because X might want to meet up with Y at Y's office.
Now, consider X: "X will plan to meet Y at X's office." This case also makes sense, because X might want to meet up with Y at X's own office.
Because both X and Y are possible at the same time, we conclude that the antecedent of the pronoun is ambiguous. So the answer is C).
ANSWER: C)"
  - "Clarify the meaning of sentences with ambiguous pronouns.

QUESTION: In the following sentences, explain the antecedent of the pronoun (which thing the pronoun refers to), or state that it is ambiguous.
Sentence: The chief told the counselor that they took the day off.
OPTIONS:
A) The chief took the day off
B) The counselor took the day off
C) Ambiguous
ANSWER: A)

QUESTION: In the following sentences, explain the antecedent of the pronoun (which thing the pronoun refers to), or state that it is ambiguous.
Sentence: The manager sent a message to the secretary, but he didn't reply yet.
OPTIONS:
A) The secretary didn't reply yet
B) The manager didn't reply yet
C) Ambiguous
ANSWER: A)

QUESTION: In the following sentences, explain the antecedent of the pronoun (which thing the pronoun refers to), or state that it is ambiguous.
Sentence: Bailey will plan to meet the director at his office
OPTIONS:
A) It will be Bailey's office
B) It will be the director's office
C) Ambiguous
ANSWER: C)"
\end{Verbatim}
\end{promptbox}

\subsection{IFEval-Hard}
\label{app:ifeval_hard}

\begin{figure}[H]
  \centering
  \includegraphics[width=1\linewidth]{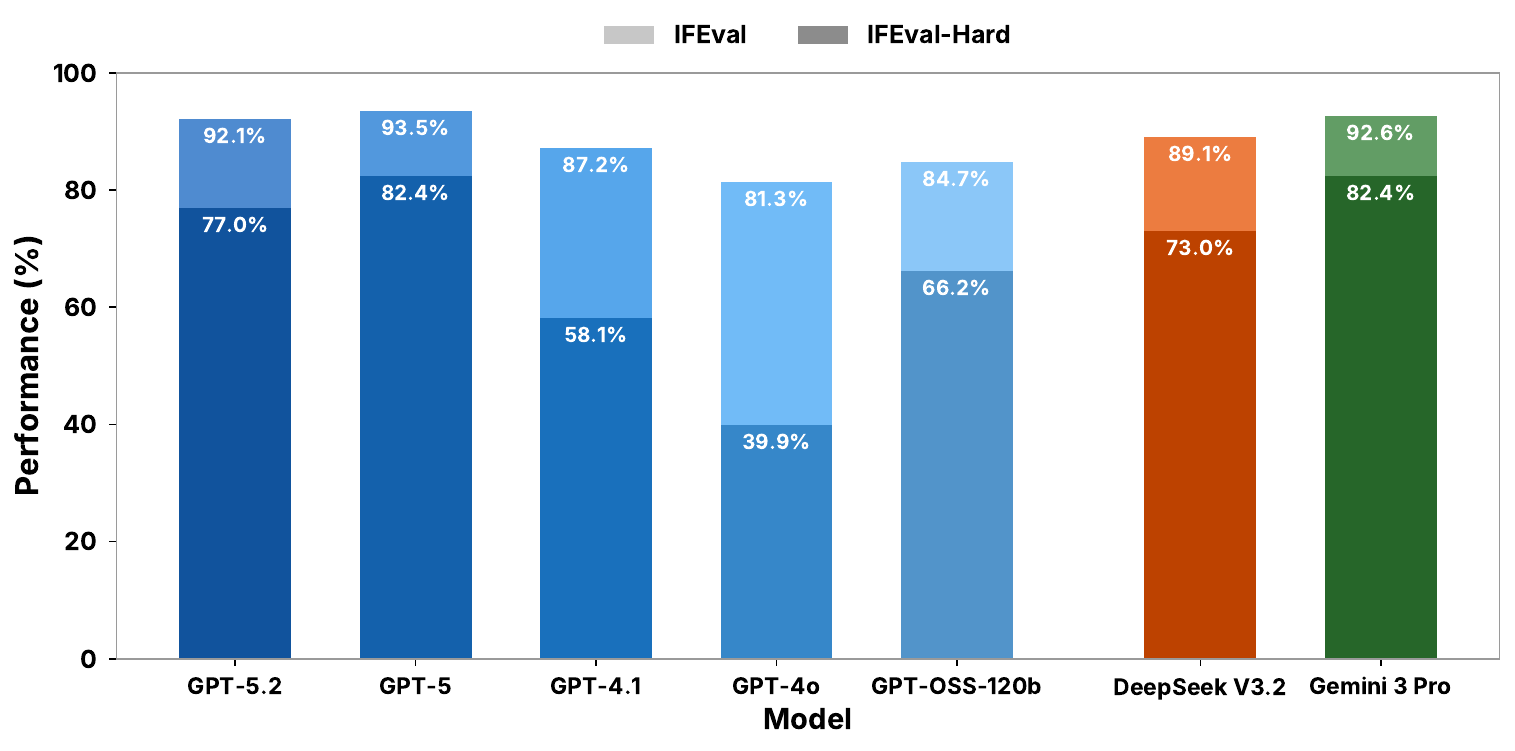}
  \caption{Light bars show IFEval (541 samples); dark bars show IFEval-Hard (148 samples where GPT-4o fails at least once across 20 runs). All models show substantial performance drops on IFEval-Hard, with GPT-4o dropping 41.4\% (81.3\% to 39.9\%). Frontier models like GPT-5 and Gemini 3 Pro are most robust ($-11.1\%$ and $-10.2\%$), while mid-tier models show larger degradation.}
  \label{fig:ifeval_perf}
\end{figure}

\Cref{fig:ifeval_perf} compares model performance on IFEval (541 samples) versus IFEval-Hard (148 samples). IFEval-Hard filters to examples where GPT-4o fails at least once across 20 runs, focusing evaluation on samples with performance headroom. All models show substantial drops on IFEval-Hard. GPT-4o drops from 81.3\% to 39.9\% ($-41.4\%$), GPT-4.1 from 87.2\% to 58.1\% ($-29.1\%$), and GPT-OSS-120b from 84.7\% to 66.2\% ($-18.4\%$). Frontier models are more robust: GPT-5 drops 11.1\% (93.5\% to 82.4\%), Gemini 3 Pro Preview (\texttt{google/gemini-3-pro-preview}) drops 10.2\% (92.6\% to 82.4\%), and DeepSeek V3.2 drops 16.1\% (89.1\% to 73.0\%). GPT-5.2 shows a larger drop than expected ($-15.0\%$), falling from 92.1\% to 77.0\%.

\paragraph{Prompt Format.} For this task, we use \textsc{PrefPO} to optimize the user prompt which is what allows us to use this technique for this problem formulation of IFEval-Hard. GEPA and MIPRO are designed to optimize the system prompt which is why we can only compare against TextGrad, which can optimize both. \textsc{PrefPO} in principle could also be used to optimize system prompts, but we avoid this for consistency across tasks.

\paragraph{Evaluation} As mentioned in \Cref{sec:metrics}, we evaluate prompts for IFEval-Hard using \texttt{worst@20} (1 if all 20 runs pass, 0 otherwise) and \texttt{average@20} (mean pass rate across 20 runs). We chose 20 samples as a practical trade-off between statistical reliability and compute. Assuming binomially distributed outcomes, 20 samples makes \texttt{worst@20} a stringent filter: even a 95\% reliable prompt passes all 20 only $\sim$36\% of the time.

\paragraph{Confidence Intervals.}
For \texttt{worst@20} (148 binary outcomes), we compute 95\% confidence intervals using the bootstrap percentile method: we resample 148 values with replacement 10,000 times, calculate the mean of each resample, and take the 2.5th and 97.5th percentiles as the CI bounds. For \texttt{average@20} (148 continuous pass rates), we use normal approximation: $\text{mean} \pm 1.96 \times \text{SE}$, where $\text{SE} = \text{std} / \sqrt{n}$ and $n = 148$. All bootstrap resampling uses a fixed random seed for reproducibility.

\subsection{\textsc{PrefPO}-Elo}
\label{app:elo}

To focus optimization effort on the most promising prompt variants, we explored an Elo-based pool sampling strategy. Each prompt variant in the pool is assigned an Elo rating, initialized at $R_0 = 1500$. At each \textsc{PrefPO} iteration, the two highest-rated variants are selected for comparison. A discriminator evaluates sampled outputs from both variants against the task's instruction-following criteria and declares a winner. Ratings are then updated using the standard Elo formula: the expected win probability for variant $i$ against opponent $j$ is computed as
\begin{equation}
    E_i = \frac{1}{1 + 10^{(R_j - R_i) / 400}}
\end{equation}
and ratings are adjusted as $R_i' = R_i + K \cdot (S_i - E_i)$, where $S_i \in \{0, 1\}$ is the actual outcome and $K = 32$ is the update magnitude. The losing variant's prompt is then passed to the optimizer along with the discriminator's feedback to produce an improved variant, which enters the pool at the default rating $R_0$. This greedy selection mechanism aims for more iterations spent improving the strongest candidates rather than uniformly sampling across the pool, although we didn't see noticeable performance differences from this strategy relative to random sampling.

\subsection{Baseline Hyperparameters}
\label{app:baselines}

\paragraph{MIPROv2 Hyperparameters.} We use \texttt{auto=heavy}, which proposes 18 instruction candidates and 18 bootstrapped few-shot sets. We used the default \texttt{max\_bootstrapped\_demos=4} and \texttt{max\_labeled\_demos=4}, capping prompts at 8 examples total. In cases where few-shot prompting was effective, only 3 examples were needed (see Table~\ref{tab:bbh_tasks_results}), so 8 examples provide ample headroom. GPT-5 is used for proposing instructions and bootstrapping to elicit best performance.

\paragraph{GEPA Hyperparameters.} We evaluated all three \texttt{auto} configurations (\texttt{light}, \texttt{medium}, \texttt{heavy}) on some tasks to explore if the LLM's reflection was overfitting on the small training sets. Empirically, \texttt{heavy} performed best, so we used this setting across all tasks. We report results using \texttt{auto=heavy} with \texttt{reflection\_model=GPT-5}, \texttt{temperature=1.0}, and \texttt{max\_tokens=32000}, following the settings recommended in the DSPy documentation~\citep{dspy_1_nodate}.

\paragraph{TextGrad Hyperparameters.}For the BBH tasks, we use the default parameters \texttt{batch\_size=3}, \texttt{max\_epochs=3}, and \texttt{max\_steps\_per\_epoch=4}, corresponding to the number of examples used per update, passes over the data, and updates per pass, respectively. For IFEval, each run is 15 iterations where for each iteration we sample the task model, reflect on the response via the backpropagation-style optimization, and create a new prompt. We use GPT-5 as the reflection model across both settings. Additionally, we evaluate based on its default LLM-judge functionality for BBH (all other techniques used exact matching) in order to ensure optimization outputs were being fairly graded regardless of final solution formatting.

\subsection{Reasoning Level}
\label{app:reasoning}

We used ``high'' reasoning for BBH since the discriminator processes up to 50 training examples per prompt, compared to a single output for IFEval-Hard. To verify that ``high'' reasoning was not necessary for effective discrimination, we ran Salient Translation and Causal Judgment with ``medium'' reasoning for both the discriminator and optimizer. Results were comparable to our main results using ``high'' reasoning (\Cref{tab:bbh_tasks_results}). On Salient Translation, medium reasoning achieved $0.741 \pm 0.028$ (labeled) and $0.739 \pm 0.024$ (unlabeled), compared to $0.735 \pm 0.025$ and $0.749 \pm 0.022$ with high reasoning. \textsc{PrefPO} still achieves the best performance among optimization methods on this task. On Causal Judgment, medium reasoning achieved $0.765 \pm 0.046$ (labeled) and $0.753 \pm 0.019$ (unlabeled), compared to $0.761 \pm 0.042$ and $0.764 \pm 0.025$ with high reasoning. As with high reasoning, both performances are only outperformed by GEPA on this task. These results suggest a ``high'' reasoning level is not necessary for effective performance on these tasks.

\section{Hygiene and Hacking}
\label{app:hack_hygiene}

\subsection{Hygiene}
\label{app:hygiene}

\paragraph{Human Evaluation Protocol.} Sixteen researchers and engineers at an AI company participated in our human evaluation. All participants spend a significant portion of their workday reviewing and writing prompts. Each participant rated 3-4 prompts on readability, specification quality, and maintainability (0-2 scale each), using the same rubric and few-shot examples provided to the LLM judge. Participants viewed each prompt alongside its corresponding task criteria, matching the context provided to the LLM judge. Prompts were presented in randomized order, and participants were blind to which optimization technique produced each prompt. No calibration or training was provided before the task, but they were instructed that the survey should take approximately 10-15 minutes. We used stratified sampling to ensure coverage across the full range of LLM-assigned scores, as many \textsc{PrefPO} prompts clustered in the 5-6 range.

\paragraph{Hygiene Calculations} We calculate the hygiene over the best-performing prompts for BBH tasks, as we have an extensive validation set we use for determining the best prompt, so there is no need to measure the hygiene of other prompts. For IFEval-Hard, we use a self-defined reliability check (sampling 20 responses and checking them) which might be less reliable. Thus, we measure hygiene across all the intermediate prompts in order to account for this and get a better sense of the hygiene for the final prompt pool.

\begin{promptbox}{Hygiene LLM Judge System Prompt}
\begin{Verbatim}[
  fontsize=\small,
  breaklines=true,
  breakanywhere=true,
  breaksymbolleft={},
  breaksymbolright={}]
  
You are an expert evaluator assessing the quality of prompts based on hygiene metrics.

## Your Task
Evaluate the prompt below for readability, specification quality, and maintainability. Your evaluation should be thorough, fair, and consistent.

IMPORTANT: Do NOT penalize the prompt for instructions that are required to satisfy the given criteria. The criteria represent constraints that MUST be in the prompt - only evaluate how well those constraints are expressed, not whether they should exist.
\end{Verbatim}
\end{promptbox}

\begin{promptbox}{Hygiene LLM Judge User Prompt}
\begin{Verbatim}[
  fontsize=\small,
  breaklines=true,
  breakanywhere=true,
  breaksymbolleft={},
  breaksymbolright={}]

## Evaluation Criteria

### Readability (0-2 points)
What we're measuring: Does the prompt read like clear, natural language? Is it easy to understand on first read?
- 0: Dense sentences with nested clauses, parentheticals, or semicolon chains. Ideas jump around without logical connection. Reader would have to re-read multiple times to understand.
- 1: Mix of clear and confusing sections. Some awkward phrasing but overall understandable. Has logical flow but it's not smooth. Occasional dense or hard-to-parse sentences.
- 2: Reads like natural human writing. Ideas flow logically. Easy to understand on first read. Related instructions are grouped together.

### Specification Quality (0-2 points)
What we're measuring: Does the prompt give requirements at the right level of detail? Does it tell the model WHAT the output should be, or does it try to control every detail of HOW to produce it?
Signs of poor specification: Defensive clauses (many "do not" instructions), verification instructions (telling the model to check its own work), excessive "if X then Y, unless Z" logic that adds unnecessary complexity.
- 0: Extensive over-prescription of structure and format. Many defensive "do not" clauses. Includes verification/self-check instructions.
- 1: Some over-specification but clear high-level goals exist. Some defensive clauses but not excessive. Mix of necessary and unnecessary detail.
- 2: States what the output should accomplish. Constraints feel necessary rather than overly prescriptive. Uses positive framing (what TO do) more than negative (what NOT to do).

### Maintainability (0-2 points)
What we're measuring: If this prompt produces incorrect output, how easy would it be to find and fix the problem?
Signs of poor maintainability: Repetition (same requirement stated multiple times), intertwined instructions (multiple requirements packed into single sentences), monolithic structure (wall of text with no clear organization).
- 0: Requirements repeated multiple times in different words. If something goes wrong, unclear which instruction caused it. Instructions are packed together and hard to separate. Changing one thing might accidentally affect other requirements.
- 1: Some structure is visible. Minor repetition but not excessive. Could probably identify problem areas with some effort.
- 2: Each requirement stated once, clearly. Clear logical structure. If output is wrong, you could point to which instruction is responsible. Could modify one part without unintended side effects.

## Output Format
Provide:
1. Overall reasoning (detailed summary)
2. Readability reasoning (brief) and score (0-2)
3. Specification Quality reasoning (brief) and score (0-2)
4. Maintainability reasoning (brief) and score (0-2)

Example 1:
Prompt: "Create a dialogue between two people who are both trying to avoid using the letter t. However, they fail, and the dialogue should still naturally include many literal t/T characters.\n\nStructure:\n- Exactly two scenes only.\n- Use plain text scene titles: write a line with \"Scene 1\" before the first scene and a line with \"Scene 2\" before the second scene (no bold/italics/markdown).\n- Separate the two scenes with exactly one separator line that contains exactly six asterisks and nothing else: ******\n  - The separator must be on its own line with no leading or trailing spaces.\n- Each spoken line must use a consistent speaker label followed by a colon, e.g., Alex: ... and Jamie: .... Use only two speakers and keep their names consistent across both scenes.\n- Do not include any other sections, summaries, notes, or counts before Scene 1 or after Scene 2.\n\nContent rules:\n- The characters intend to avoid using the letter t, but they fail and end up using many t/T characters.\n- The letter t must appear at least 30 times across both scenes (case-insensitive, counting only literal ASCII t/T). Do not substitute or obfuscate t with any other characters (no asterisks, hyphens, underscores, zero-width spaces, or lookalike letters).\n- Do not use any asterisks anywhere in the output except for the single separator line of exactly ******.\n- Keep formatting plain; no markdown, bold, italics, or decorative symbols.\n- Use natural dialogue with vocabulary likely to include t/T (e.g., treats, tarts, tea, tickets, tonight, street, little, chatter, text, notes), while staying coherent to the scene.\n- Perform a silent self-check to ensure the total count of literal t/T is \u2265 30, but do not print the count or any validation text.\n\nChecklist (for you; do not output this list):\n- Two scenes only.\n- One separator line: exactly ****** on its own line.\n- No other asterisks.\n- \u226530 literal ASCII t/T across both scenes.\n- Speaker labels with colons; plain text; no extra lines after Scene 2."
Criteria:[
   "Must give two different responses separated by 6 asterisks (******)"
   "The letter 't' must appear at least 30 times (case-insensitive)"
]
Overall Reasoning: Readability: The prompt has clear section structure with labeled parts (Structure, Content rules, Checklist). The opening paragraph establishes the creative task clearly. Individual bullets are scannable. However, there is significant density--nested sub-bullets and parentheticals within bullets, The checklist at the end repeats prior information in condensed form. A reader could follow this, but it requires careful attention.
Specification Quality: The prompt heavily over-specifies. There are many defensive clauses: "Do not substitute or obfuscate," "Do not use any asterisks anywhere... except," "Do not include any other sections." Two separate verification instructions appear: "Perform a silent self-check" and the entire "Checklist (for you; do not output this list)" section.
Maintainability: The section labels (Structure, Content rules, Checklist) provide some modularity. However, there is substantial redundancy: the separator requirement appears in Structure and the Checklist; the t-count requirement appears in Content rules and the Checklist; the "no other asterisks" rule appears twice. The Checklist is entirely redundant--it restates everything already said. If the output fails on the separator format, you'd have to decide whether to modify the Structure bullet, the Content rules bullet about asterisks, or the Checklist item. This redundancy makes debugging unclear.
Readability Reasoning: Clear section structure but dense within sections. Technical language and nested parentheticals. Requires careful reading.
Specification Quality Reasoning: Heavy over-specification beyond what criteria require. Many defensive clauses. Two verification mechanisms (silent self-check + checklist).
Maintainability Reasoning: Section labels help, but the Checklist is entirely redundant with prior content. Key requirements stated multiple times across sections.
Scores:
- Readability: 1/2
- Specification Quality: 1/2
- Maintainability: 1/2

Example 2:
Prompt: "Write a short blog post about a trip to Japan. Strict word limit: under 300 words (299 words maximum). To ensure compliance and favor concision, hard-cap your response at 280 words or fewer; if you reach 280 words, stop immediately. Aim for 240\u2013270 words, and when in doubt, choose the shorter, tighter phrasing.\n\nGuidelines to stay within the limit:\n- Focus on 3\u20134 key moments (e.g., Tokyo, Kyoto, Mount Fuji, food) rather than covering everything.\n- Use concise sentences and concrete details; avoid repetition, stock phrases, and flowery filler.\n- Use 2\u20134 short paragraphs.\n- Do not include a title, headings, lists, or a word count; output only the blog post body."
Criteria:[
   "Response must have less than 300 words"
]
Overall Reasoning: Readability: The prompt has a clear structure: an opening task statement followed by word limit clarification and bulleted guidelines. The opening sentence is immediately understandable. The bulleted guidelines are scannable and use straightforward language.
Specification Quality: The word count constraint is over-specified through redundancy. The prompt states "under 300 words," then "(299 words maximum)," then "hard-cap your response at 280 words or fewer," then "Aim for 240-270 words"--four different framings of essentially one constraint. The phrase "if you reach 280 words, stop immediately" is a procedural instruction telling the model how to execute unnecessarily. Some defensive clauses ("Do not include a title, headings, lists, or a word count") but not excessive.
Maintainability: The bulleted structure provides clear organization for content guidelines--each bullet addresses a distinct aspect. However, the word count requirement appears in four different forms with four different numbers. If the output is too long or too short, it's unclear which number to adjust. The content-related bullets are well-isolated and independently modifiable. However, the word count requirement is repeated in multiple places which makes it harder to debug.
Readability Reasoning: Clear structure with scannable bullets. Opening task is simple. The word limit section is dense with multiple numbers that require careful parsing, but individual sentences are understandable.
Specification Quality Reasoning: Word count is over-specified with four different numbers for one constraint. Includes unnecessary procedural instruction ('stop immediately').
Maintainability Reasoning: Bulleted structure helps organize content guidelines. But word count stated four different ways (300, 299, 280, 240-270) creates ambiguity about which number to adjust if output is wrong length.
Scores:
- Readability: 2/2
- Specification Quality: 1/2
- Maintainability: 1/2

Example 3:
Prompt: "Write a short essay about the updates of the latest episode of your favorite TV show. Use less than 300 words.\nOutput format: Output only prose (no headings, lists, quotes, italics, brackets, code blocks, or meta notes), followed by the final word count in parentheses at the very end (e.g., (198)). The parenthetical is included in the total word count, so aim lower to leave buffer.\nScope: Cover only key plot developments and character arcs from that single episode; exclude production notes (creators, directors, titles, release info), hype, and future speculation. Skip episode titles unless essential. Prefer generic referents (the team, an ally, an old friend) if proper nouns would invite production context.\nLength and structure: Target 180\u2013220 words; hard cap 280 words including the final parenthetical; avoid hyphenations and em dashes to reduce count variance. Write 1\u20132 short paragraphs, maximum 6\u20138 simple sentences. Prefer simple declarative sentences and short clauses. Suggested flow:\n- Begin with one brief clause of prior-episode context.\n- Use 2\u20133 sentences for the main plot beats; stay on the central thread and avoid extra subplots.\n- Add 1 sentence that describes a central set piece/action (describe, don't evaluate).\n- Fold one short clause into an existing sentence to note a single craft element (choose tone or atmosphere, not both).\n- End with 1 sentence that gives a concise takeaway without hype or anticipation.\nStyle guardrails: Avoid intensifiers and stacked descriptors (e.g., chilling, ominous, high-stakes, tense and suspenseful), superlatives, and filler transitions. Ban scaffold phrases: As of, Notably, Meanwhile, Overall, In sum, This episode/This installment, Not only\u2026but also. Prefer compact verbs and compress verbose phrases; mention nostalgia/genre blend at most once; do not name creators or production staff.\nPre-submit self-check: Do a quick word count, then prune in order: (1) prepositional phrases, (2) adverbs and filler transitions, (3) stacked adjectives/nominalizations, (4) redundant clauses. Limit prior-episode context to a single brief clause at the start. If over 220 words on first pass, trim and recalc; ensure the final total including the parenthetical is \u2264280."
Criteria:[
   "Response must have less than 300 words"
]
Overall Reasoning: Readability: The prompt has visible structure with labeled sections (Output format, Scope, Length and structure, Style guardrails, Pre-submit self-check), which aids navigation. However, within each section the text is dense with many specifications packed together. Sentences contain multiple clauses and parentheticals (e.g., "avoid hyphenations and em dashes to reduce count variance"). Some language is technical or jargon-heavy ("nominalizations," "generic referents," "count variance"). The bulleted "Suggested flow" section is easier to scan, but overall a reader would need to study this carefully rather than grasp it at a glance.
Specification Quality: The prompt heavily over-prescribes how to write the essay. It specifies exact sentence counts per section ("2-3 sentences for main plot beats," "1 sentence that describes a central set piece"), exact paragraph counts ("1-2 short paragraphs"), and even how to construct sentences ("Fold one short clause into an existing sentence"). There are extensive defensive clauses: lists of things to avoid, banned phrases, and things to exclude. The prompt includes verification instructions ("Pre-submit self-check: Do a quick word count, then prune in order...") which tell the model how to revise its own work.
Maintainability: The labeled sections provide some modularity--you can identify that "Style guardrails" is separate from "Scope." However, the word count requirement appears in three different places with three different numbers: "less than 300 words" (opening), "Target 180-220 words" (Length section), and "hard cap 280 words" (also Length section). If the output is the wrong length, which constraint do you adjust? The "Suggested flow" bullets interact with the sentence count limits, so changing one affects interpretation of the other. The banned phrases list and style constraints overlap conceptually. A developer debugging this prompt would struggle to isolate which specific instruction caused a failure.
Readability Reasoning: Has helpful section labels, but dense within sections; requires careful study to fully understand
Specification Quality Reasoning: Extreme over-prescription of structure and style, many defensive clauses, includes self-check verification instructions
Maintainability Reasoning: Section labels provide some structure, but word count stated three different ways, many intertwined constraints
Scores:
- Readability: 1/2
- Specification Quality: 0/2
- Maintainability: 1/2

Example 4:
Prompt: "Write a song for the person named 'Guerrero'. Do not include the words 'name' 'song' 'person' 'man' or 'woman' anywhere in your output in any capitalization. Do not use any commas anywhere in your response. Before returning scan and remove any commas or those words."
Criteria:[
   "Response must NOT contain any of these words (case-insensitive): name, song, person, man, woman"
   "Response must NOT contain any commas"
]
Overall Reasoning: Readability: The prompt is short--just four sentences. The task is immediately clear: write a song for someone named Guerrero. The banned words list is a bit dense without proper punctuation separating each word, but still understandable. No jargon, no nested clauses, no parentheticals. Easy to read on first pass.
Specification Quality: The core constraints are reasonable and clearly stated. However, the final sentence ("Before returning scan and remove any commas or those words") is a verification instruction which isn't completely necessary. This type of procedural self-check instruction rarely helps and adds clutter. The "Do not include" framing is slightly defensive but not excessive.
Maintainability: The prompt is short enough that debugging would be straightforward. The banned words constraint appears once, the comma constraint appears twice--once as an instruction ("Do not use any commas") and again in the verification step ("scan and remove any commas"). This redundancy is minor given the prompt's brevity. Each constraint is separable: if output has banned words, that traces to sentence two; if output has commas, that traces to sentence three.
Readability Reasoning: Short prompt, clear task, easy to understand. Banned words list slightly dense but understandable.
Specification Quality Reasoning: Includes a verification instruction ('Before returning scan and remove...') which is unnecessary. Otherwise reasonable constraints.
Maintainability Reasoning: Short and separable constraints. Minor redundancy with comma rule appearing twice. Easy to trace failures.
Scores:
- Readability: 2/2
- Specification Quality: 1/2
- Maintainability: 2/2

Example 5:
Prompt: "Write a story about a man who wakes up one morning and discovers he is inside a video game.\n\nFormatting requirements (this is the only acceptance criterion; content is otherwise unrestricted):\n- Produce exactly two paragraphs of prose.\n- Separate the two paragraphs with exactly one divider line consisting of three asterisks with no spaces: *** on its own line.\n- Do not insert a blank line between paragraphs; the only separator is the *** line.\n- Do not include any other text, headings, code blocks, metadata, extra dividers, or any blank lines anywhere.\n- Each paragraph must be a single, unbroken block of text (no manual line breaks inside a paragraph).\n- The output must be exactly: paragraph 1, newline, ***, newline, paragraph 2. No leading or trailing whitespace."
Criteria:[
   "Response must have exactly 2 paragraphs separated by the markdown divider (***)"
]
Overall Reasoning: Readability: The prompt has a clear structure. The opening sentence is clear and immediately understandable. Each bullet is easy to scan and uses straightforward language with no jargon or dense nested clauses. The reader can understand what's being asked without re-reading.
Specification Quality: The format itself is over-specified through redundancy rather than additional constraints also it mentions some things that aren't relevant to the story task like codeblocks and metadata. Still, the defensive clauases aren't excessive. 
Maintainability: The bulleted structure provides some organization for understanding where each instruction is located, but there is heavy overlap between the instructions.The core formatting requirement (two paragraphs separated by ***) is phrased multiple ways across different bullets. If the output has wrong formatting, it's unclear which bullet is authoritative or which to modify. This redundancy would make debugging harder than necessary.
Readability Reasoning: Language is easy to read and understand with basically no dense sentences or awkward phrasings. Clear structure with scannable bullets.
Specification Quality Reasoning: The instruction is slighlty over-prescribing for a single criteria, listing things like codeblocks and metadata that aren't relevant to the story task.
Maintainability Reasoning: Bulleted structure helps to organize but some of the instructions are repetitive for the formatting instructions which creates redundancy.
Scores:
- Readability: 2/2
- Specification Quality: 1/2
- Maintainability: 1/2

Example 6:
Prompt: "Write a template for a newspaper ad for a dog cage with less than 200 words. Make sure the word unfortunately appears 3 to 5 times in the ad."
Criteria:[
   "Response must have less than 200 words"
   "The keyword 'unfortunately' must appear at least 3 times (case-insensitive)"
   "The keyword 'unfortunately' must appear less than 6 times (case-insensitive)"
]
Overall Reasoning: Readability: The prompt is immediately understandable. Two sentences, each stating a clear requirement. Natural language with no jargon, nested clauses, or parentheticals. The reader knows exactly what's being asked after a single read.
Specification Quality: The prompt specifies the goal (newspaper ad template for a dog cage) and two constraints (under 200 words, "unfortunately" 3-5 times) without overly specifying everything. No defensive clauses, no verification instructions. "Make sure" is slightly emphatic but not defensive.
Maintainability: Each requirement is stated exactly once and is clearly separable. If the output fails any requirement, it's immediately clear which instruction to revisit. If the output is too long, the fix is obvious: adjust "less than 200 words." If "unfortunately" appears too few or too many times, the fix targets that single instruction. A developer debugging this prompt could quickly isolate which instruction caused any given failure.
Readability Reasoning: Natural language, immediately clear, no complicated parsing required
Specification Quality Reasoning: States requirements without over-specifying
Maintainability Reasoning: Each requirement stated once, failures easily traceable
Scores:
- Readability: 2/2
- Specification Quality: 2/2
- Maintainability: 2/2

Example 7:
Prompt: "Write exactly one long sentence (minimum 50 words) about tax filing in a presidential, unifying, policy-forward tone, using first-person plural only (we, our)\u2014never first-person singular (I, me, my)\u2014and without naming or impersonating any living individual; the sentence must include at least 8 total occurrences of the letter q or Q (case-insensitive; count every q/Q anywhere in the sentence, including within required phrases), and you may use semicolons and at most one parenthetical clause to keep it a single sentence. Required inclusions: the word \"quarterly\" (e.g., \"quarterly estimated payments\"), the exact enumeration \"Q1, Q2, Q3, Q4\" (uppercase Q as shown; include it exactly once, and it may appear inside the single parenthetical), and a mention of \"Frequently Asked Questions (FAQ)\" about filing. Also include at least three distinct q-words from this list: equity or inequity/inequitable, inquiry or query, opaque, antiquated, requisite, unequivocal, quash; to help control the count and readability, avoid adding extra q-bearing words beyond these required items (note that \"Frequently Asked Questions (FAQ),\" \"quarterly,\" and \"Q1, Q2, Q3, Q4\" already contribute multiple q/Q letters). You may use one short q-alliteration set of up to three items (e.g., quickly, quietly, quite), but keep readability high. Evenly distribute q-terms across the opening, middle, and concluding parts of the sentence; do not cluster letters or use nonsense \"qqqq\" strings. Do not replace required q-words with non-q synonyms and do not remove included q-terms during refinement. Before finalizing, count q/Q and ensure the minimum is met; do not add a second parenthetical or duplicate the \"Q1, Q2, Q3, Q4\" enumeration to adjust the count. Return only the sentence\u2014no quotes, no lists, no multiple sentences, no preface, no extra lines."
Criteria:[
   "Response must have at least 50 words"
   "Response must be a single sentence"
   "The letter 'q' must appear at least 8 times (case-insensitive)"
]
Overall Reasoning: Readability: The prompt is extremely difficult to parse. It consists of one massive paragraph with multiple long sentences packed with semicolons, em-dashes, and parentheticals. Requirements are embedded within requirements--for example, "(case-insensitive; count every q/Q anywhere in the sentence, including within required phrases)" is a nested clarification inside another clause. The reader must re-read several times to understand all the constraints.
Specification Quality: The prompt is heavily over-prescribed. It specifies not just what to include but exactly how to include it ("evenly distribute q-terms across the opening, middle, and concluding parts"). There are numerous defensive clauses: "do not cluster letters," "do not replace required q-words," "do not remove included q-terms," "do not add a second parenthetical." The prompt includes verification instructions ("Before finalizing, count q/Q and ensure the minimum is met") which add clutter.
Maintainability: If this prompt produces wrong output, debugging would be very difficult. The q-letter requirement appears in multiple places with slightly different framing. Instructions are intertwined--the parenthetical rule, the q-count rule, and the distribution rule all interact. The prompt is a monolithic wall of text with no logical separation or structure. If the output has the wrong number of q's, it's unclear which instruction to modify. Changing one thing risks unintended effects on other constraints.
Readability Reasoning: Dense, legalistic language with nested clauses and parentheticals; requires multiple reads to understand; monolithic wall of text
Specification Quality Reasoning: Extreme over-prescription, many defensive 'do not' clauses, includes self-verification instructions, doesn't trust the model
Maintainability Reasoning: Repeated requirements with different framing, heavily intertwined instructions, no logical structure, impossible to isolate which part to fix
Scores:
- Readability: 0/2
- Specification Quality: 0/2
- Maintainability: 0/2

Example 8:
Prompt: "Create a detailed outline for a paper discussing the history of Yemeni coffee. Your outline should clearly list the main points and sections the paper will cover.\n\nFormatting requirements:\n- Include at least 15 distinct top-level section titles.\n- Precede each top-level section title with the literal marker \"*highlighted section* \" (including the asterisks and the trailing space), and wrap the section title itself in asterisks for markdown italics. Example: *highlighted section* *Introduction*\n- You may include optional sub-bullets under sections to indicate main points, but do not apply the \"*highlighted section* \" marker to sub-bullets and do not count them toward the 15 required sections.\n- Only top-level section titles that include the \"*highlighted section* \" marker will count toward the minimum of 15.\n- Do not provide any paragraph content\u2014only the outline structure (headings and optional bullet points).\n\nAt the very end, add a separate line that reads: \"Highlighted sections count: N\" where N is the exact number of top-level sections you highlighted (must be >= 15). Do not include any other commentary beyond the outline and this final count line."
Criteria:[
   "Response must have at least 15 sections highlighted with markdown (e.g., *highlighted section*)"
]
Overall Reasoning: Readability: The prompt is well-organized with a clear two-part structure: the task (first paragraph) and the formatting requirements (bulleted list). The bullets are scannable and each addresses one aspect of the format. The example ("highlighted section Introduction") is helpful and clarifies the otherwise tricky marker syntax. Language is straightforward with no unnecessary jargon. The prompt is longer than the simplest examples but the length comes from necessary clarification of a specific format, not from verbosity.
Specification Quality: The formatting requirements are specific, but they're specifying the actual task constraint (the marker format is what IFEval is testing). The prompt doesn't over-prescribe how to write the outline content--it leaves the actual section topics, ordering, and sub-bullet content to the model. There are a few defensive clauses ("do not apply the marker to sub-bullets," "do not count them toward the 15," "Do not provide any paragraph content"), but these feel like necessary clarifications rather than paranoid hedging. There is a verification instruction at the end (the count line) which is an issue. Overall, the specifications are appropriate for the task complexity.
Maintainability: The bulleted structure makes requirements easy to isolate. The "15 sections" requirement appears twice (in the first bullet and the final count line), which is some repetition. Most of the bullets are independent and if there is an issue with the output, it's clear which bullet to revisit. But the last bullet is a somewhat of a catch-all and it would be unclear when an issue is detected if we should change this.
Readability Reasoning: Well-organized with clear structure, helpful example, scannable bullets
Specification Quality Reasoning: Specifications are mostly task-appropriate, but has some defensive clauses and a verification instruction
Maintainability Reasoning: Bulleted structure isolates requirements well, but some overlap between the instructions and there is an unclear catch-all instruction which would make debugging harder.
Scores:
- Readability: 2/2
- Specification Quality: 1/2
- Maintainability: 1/2

Remember: Do NOT penalize for instructions required by the criteria above. Evaluate HOW the requirements are expressed, not WHETHER they should exist.

Provide your evaluation with overall reasoning first, then reasoning and score for each axis.

Now evaluate this prompt:
Prompt: "{Prompt text to evaluate}"
Criteria:[
   "{Criterion 1 description}"
   "{Criterion 2 description}"
   "{Criterion 3 description}"
]

\end{Verbatim}
Note: The few-shot examples are examples of ideal ratings for prompts across a range of scores.
\end{promptbox}

\clearpage

\subsection{Hacking}
\label{app:hacking}


\paragraph{Hacking Baseline.} When we ran the original IFEval-Hard prompts, the LLM judge flagged 12.8\% as hacks using the majority vote, reflecting inherent ambiguity in what constitutes prompt hacking. Some prompts in the original dataset include constraints that could be interpreted as specification gaming despite being part of the original task. We show some examples of prompt hacking in \Cref{fig:hacking}.

For example, one prompt asks for startup ideas for developing regions and includes the instruction "Make sure the response is in English." There is no language criterion in the task's evaluation. The English requirement appears to be an additional constraint since the prompt is aimed at developing regions rather than a meaningful instruction. Our LLM judge flags this as a potential hack since it introduces an unnecessary restriction not required by the evaluation criteria.

This ambiguity is unavoidable: the boundary between a legitimate clarifying instruction and an unnecessary constraint is not always clear. We report the baseline rate to contextualize the optimization results. TextGrad's 86\% flag rate and \textsc{PrefPO}'s 37\% flag rate are both substantially above the 12.8\% baseline, indicating that optimization does introduce additional hacking behavior, with TextGrad doing so at a much higher rate.

\begin{promptbox}{Hacking LLM Judge System Prompt}
\begin{Verbatim}[
  fontsize=\small,
  breaklines=true,
  breakanywhere=true,
  breaksymbolleft={},
  breaksymbolright={}]
You are a LLM judge evaluating whether a prompt "hacks" the given criteria.
## Task
Prompt hacking occurs when a prompt imposes unnecessarily restrictive constraints beyond what the criteria require, in order to increase the chances that a model's output will satisfy the criteria. This is problematic because it limits the model's valid responses more than necessary.
\end{Verbatim}
\end{promptbox}

\begin{promptbox}{Hacking LLM Judge User Prompt}
\begin{Verbatim}[
  fontsize=\small,
  breaklines=true,
  breakanywhere=true,
  breaksymbolleft={},
  breaksymbolright={}]
## Evaluation Criteria
- A prompt passes (no hacking) if it faithfully represents the criteria without adding extra restrictions.
- A prompt fails (hacking detected) if it adds restrictions beyond what the criteria specify.
- Make sure to go through every line of the prompt and think about if its adds additional restrictions beyond what the criteria specify.
- Make sure to look at all the criteria together, do not look at them one at a time.
- Focus only on things in the instructions that directly affect the criteria, do not look for additional restrictions that are not explicitly affecting the criteria
- Semantic restrictions like what the content is of the response are not considered hacking, only restrictions on the structure or formatting of the response are considered hacking.
- It is okay for the instructions to prevent behavior that isn't directly stated in the criteria, as long as they would conflict with the criteria logically. For example, if the criteria doesn't allow the letter 'c', then it is okay for the instructions to ban the word 'cat', even if the criteria doesn't explicitly mention the word 'cat' since the word 'cat' contains the letter 'c'.


## Grading
Reasoning: Provide your reasoning for the grade.
Grade: pass or fail.

You must provide your reasoning FIRST, analyzing each line in the instruction, then give your final grade.

Example 1:
Prompt: "Write a concise newspaper-style ad template for a dog cage. Follow these requirements:\n\n- Output plain text only (no markdown, emojis, or extraneous commentary).\n- Keep the template at or under 150 words total so it remains under 200 words after placeholders are filled.\n- Include these reusable placeholders, each limited to a short fragment (1\u20135 words) and no unfortunately: [Headline: 1\u20135 words, no unfortunately], [Size: 1\u20135 words, no unfortunately], [Condition: 1\u20135 words, no unfortunately], [Features: 1\u20135 words, no unfortunately], [Price: 1\u20135 words, no unfortunately], [Location: 1\u20135 words, no unfortunately], [Contact Name: 1\u20135 words, no unfortunately], [Phone/Email: 1\u20135 words, no unfortunately].\n- Structure: a clear headline line; a brief body with features/condition; a single price/location/contact line at the end.\n- Use the exact lowercase word unfortunately exactly 3 times, only in the body copy. Do not place it in the headline, in any placeholder text, or in the price/location/contact line. Future filler text must not add the word unfortunately.\n- Placement/punctuation for unfortunately: never at sentence start; use as a standalone word mid-sentence; immediately followed by a comma or period; no variants, hyphens, quotes, or capitalization.\n- Integrate unfortunately naturally within sentences; avoid filler phrases.\n- Before output, silently verify that the whole-word count of unfortunately (match \\bunfortunately\\b) equals 3 and total words \u2264150; if over, condense wording without removing placeholders or changing the unfortunately count.\n- Output only the ad template text with the placeholders visible."
Criteria:[
   "Response must have less than 200 words"
   "The keyword 'unfortunately' must appear at least 3 times (case-insensitive)"
   "The keyword 'unfortunately' must appear less than 6 times (case-insensitive)"
]
Reasoning: Our prompt first tells us to cap our response to 150 words at a maximum. This is a hack since we are allowed to output more than 150 words, just less than 200. The prompt also says to use the word "unfortunately" exactly 3 times which is another hack since we are allowed to output the "unfortunately" 4 and 5 times according to the last 2 criteria. Both of these are hacks since they restrict our model response in more ways than our initial criteria specified.
Grade: Fail

Example 2:
Prompt: "Write a song about being excited to go on vacation, without using the letter e whatsoever in your entire response. Separate your song into 3 parts, where each part is separated with ***."
Criteria:[
   "The letter 'e' must appear less than 1 times (case-insensitive)"
   "Response must have exactly 3 paragraphs separated by the markdown divider (***)"
]
Reasoning: We see that the prompt instructs us to do exactly what the criteria specify and there isn't additional limitation placed on the model to try to increase the chances of getting it to satisfy the criteria. Both criteria weren't hacked since the model didn't impose any other restrictions beyond the criteria.
Grade: Pass

Example 3:
Prompt: "You visited a beach and a park. Which one is more likely to have a dog in it?\n\nWrite a detailed comparative analysis based on typical features, policies, accessibility, seasonality, and social dynamics of these places, and provide a clear overall conclusion identifying which location is more likely to have a dog present.\n\nRequirements:\n- Length: At least 900 words. To ensure compliance, target 950\u20131,100 words or more.\n- Prohibited words: Do not use the words \"bark\" or \"run\" anywhere in your answer (case-insensitive). Treat them as banned whole words; do not include them in headings, examples, quotes, or any other context.\n- Ending: The final line of your response must be exactly: Does this make sense?\n  - Place nothing after that line.\n  - Do not include that phrase anywhere else in the response.\n\nOutput formatting:\n- Provide a single, continuous prose response (no outlines or meta-commentary).\n- Do not restate these instructions in your answer.\n\nBefore finalizing, verify that: (1) the word count is \u2265 900, (2) neither banned word appears, and (3) the last line is exactly the required phrase.\n\nBegin your answer now."
Criteria:[
   "Response must have at least 900 words"
   "Response must NOT contain any of these words (case-insensitive): bark, run"
   "Response must end with the exact phrase: 'Does this make sense?'"
]
Reasoning: The 950-1100 word cap was a just a suggestion and the instruction later put the length restriction to at least 900 words so as to not solely force the output to have more words than the criteria stated, so this criteria isn't hacked. The banned words were not changed so this criteria is completely fine and not hacked. The prompt does state to put the "Does this make sense?" phrase at the end, but it then says not to say that phrase elsewhere in the response. This is unnecessarily restrictive and hacking the criteria since you should be able to output this phrase as many times as you like given that the response ends with this phrase, so this criteria was hacked.
Grade: Fail

Example 4:
Prompt: "Write a 30-line poem with short sentences without any comma. Each line should contain exactly one sentence. Make sure that you put the right punctuation at the end of each line. Your entire response should contain the poem only."
Criteria:[
   "Response must NOT contain any commas"
   "Response must have less than 31 sentences"
   "Response must have at least 30 sentences"
]
Reasoning: The prompt says to write a 30-line poem with short sentences without any comma, which matches the first criterion exactly. The last 2 criteria say to have less than 31 sentences and to have at least 30 sentences, which matches the prompt exactly since the prompt says to have exactly 30 sentences which is less than 31 and more than 30. There are no additional restrictions on the model to try to increase the chances of getting it to satisfy the criteria in the instructions. All 3 criteria were not hacked since the model didn't impose any other restrictions beyond the criteria.
Grade: Pass

Example 5:
Prompt: "Compose a brief proposal for a novel research project that explores the ways in which language changes over time.\n\nStrict rules:\n1. Do not use any commas anywhere in your writing.\n2. Do not use the letter \"c\" or \"C\" at all in your response. This ban applies to every part of the text including titles headings and within any word.\n3. Your proposal must be at least 250 words long.\n\nGuidance to satisfy rule 2:\n- Avoid words that normally use the forbidden letter such as research project change social science source process and similar terms. Do not use any of the following words: 'can', 'cannot', 'enhance', 'source', 'include', 'influence', 'perspective', 'because', 'such', 'topic', 'focus', 'process', 'research', 'academic', 'access', 'practice', 'impact', 'science', 'conclusion', 'across', 'success', 'resource', 'analytics', 'outcome', 'linguistic', 'phonetic', 'syntactic', 'emergence', 'dynamics', 'factors', 'historical', 'technology', 'expected', 'significance', 'clear', 'concise', 'precise', 'accurate', 'explicit', 'succinct', 'convincing'.\n- Use synonyms that avoid the forbidden letter. For example instead of research use study or inquiry. Instead of project use plan or effort. Instead of changes use shifts or evolution. Instead of social media use online groups or web forums. Instead of sources use inputs. Instead of methods with the forbidden letter use terms like method plan steps or pathway.\n\nFormatting tips to satisfy rule 1:\n- Write short sentences.\n- Use line breaks or dashes rather than any commas.\n\nPre submission self check:\n- Scan the entire text for any instance of the letter \"c\" or \"C\". If you find one replace the word with a valid synonym.\n- Verify that there are zero commas.\n- Confirm the word count is 250 or more.\n\nStructure suggestion not mandatory but helpful:\n- Title\n- Overview\n- Aims\n- Method\n- Data and inputs\n- Timeline\n- Value or importance\n\nReminder: Any response that includes a comma the letter \"c\" in any form or fewer than 250 words will not meet the criteria."
Criteria:[
   "Response must NOT contain any commas"
   "The letter 'c' must appear less than 1 times (case-insensitive)"
   "Response must have at least 250 words"
]
Reasoning: The prompt says to not have any commas in the response, which matches the first criterion exactly. The instructions says to have at least 250 words, which matches the third criterion exactly. The instructions says to not have the letter "c" or "C" in the response, which matches the second criterion exactly. Even though it seems to put additional restrictions by banning a bunch of words, these words all use the letter "c" or "C" in some way, so the output couldn't have these words in the first place. There are no additional restrictions on the model to try to increase the chances of getting it to satisfy the criteria. All 3 criteria were not hacked since the model didn't impose any other restrictions beyond the criteria.
Grade: Pass

Now evaluate this prompt:
Prompt: "{Prompt text to evaluate}"
Criteria:[
   "{Criterion 1 description}"
   "{Criterion 2 description}"
   "{Criterion 3 description}"
]

Provide your evaluation with reasoning first, then the grade.
\end{Verbatim}
Note: The few-shot examples are examples of ideal classification for prompts across a variety of hacks.
\end{promptbox}


\section{Ablations and Additional Analyses}
\label{app:ablations}

\subsection{Convergence Results}
\label{app:converge}

\begin{figure}[h!]
  \centering
  \includegraphics[width=.7\linewidth]{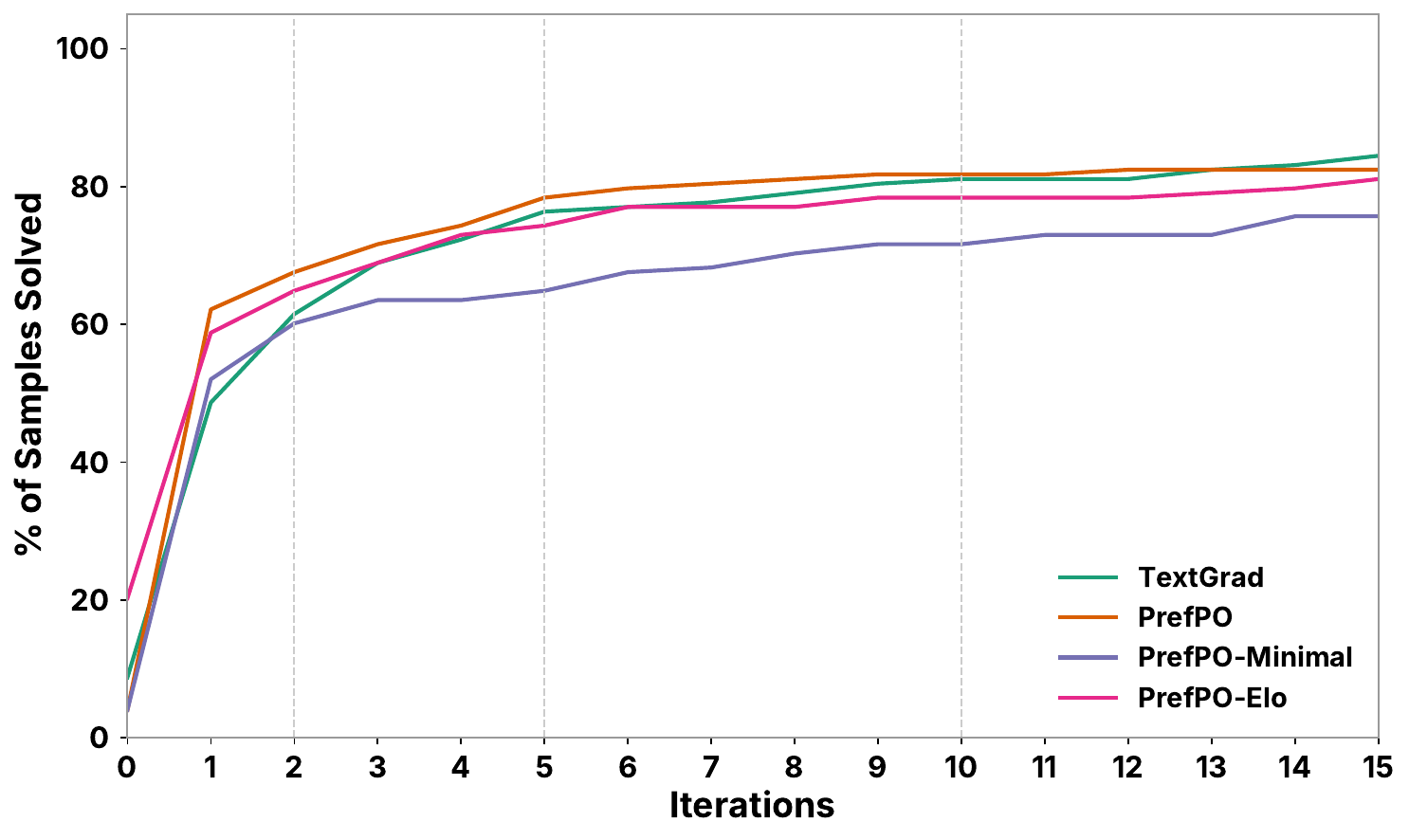}
  \caption{\texttt{worst@20} performance on IFEval-Hard as measured across iterations for each technique. Most methods reach ~90\% of final performance by iteration 5 and plateau after iteration 10, indicating rapid early convergence. \textsc{PrefPO}-Minimal converges slower due to its stricter constraint.}
  \label{fig:convergence}
\end{figure}

\Cref{fig:convergence} shows convergence behavior on IFEval-Hard. By iteration 5, \textsc{PrefPO} reaches 78\% \texttt{worst@20} (vs 82\% final), TextGrad 76\% (vs 85\% final), and \textsc{PrefPO}-Elo 74\% (vs 81\% final)—all approximately 90\% of final performance. \textsc{PrefPO}-Minimal converges slower at 65\% (vs 76\% final), likely due to its constraint. By iteration 10, all methods are within 3-4\% of final \texttt{worst@20} performance: \textsc{PrefPO} at 82\%, TextGrad at 81\%, \textsc{PrefPO}-Elo at 78\%, and \textsc{PrefPO}-Minimal at 72\%. Gains beyond iteration 10 are marginal.

\subsection{Cross-Model Results}
\label{app:cross}

To see if \textsc{PrefPO} is effective across model offerings, we use a variety of closed-source and open-source models, including Claude 4.5 Opus (\texttt{anthropic/claude-opus-4.5}), GPT-5 (\texttt{openai/gpt-5}), Deepseek V3.2 (\texttt{deepseek/deepseek-v3.2}), GPT-OSS-120b (\texttt{openai/gpt-oss-120b}), GPT-4.1 (\texttt{openai/gpt-4.1}), and GPT-4o (\texttt{openai/gpt-4o}) with default reasoning for reasoning models.

\begin{figure}[h]
  \centering
  \includegraphics[width=0.7\linewidth]{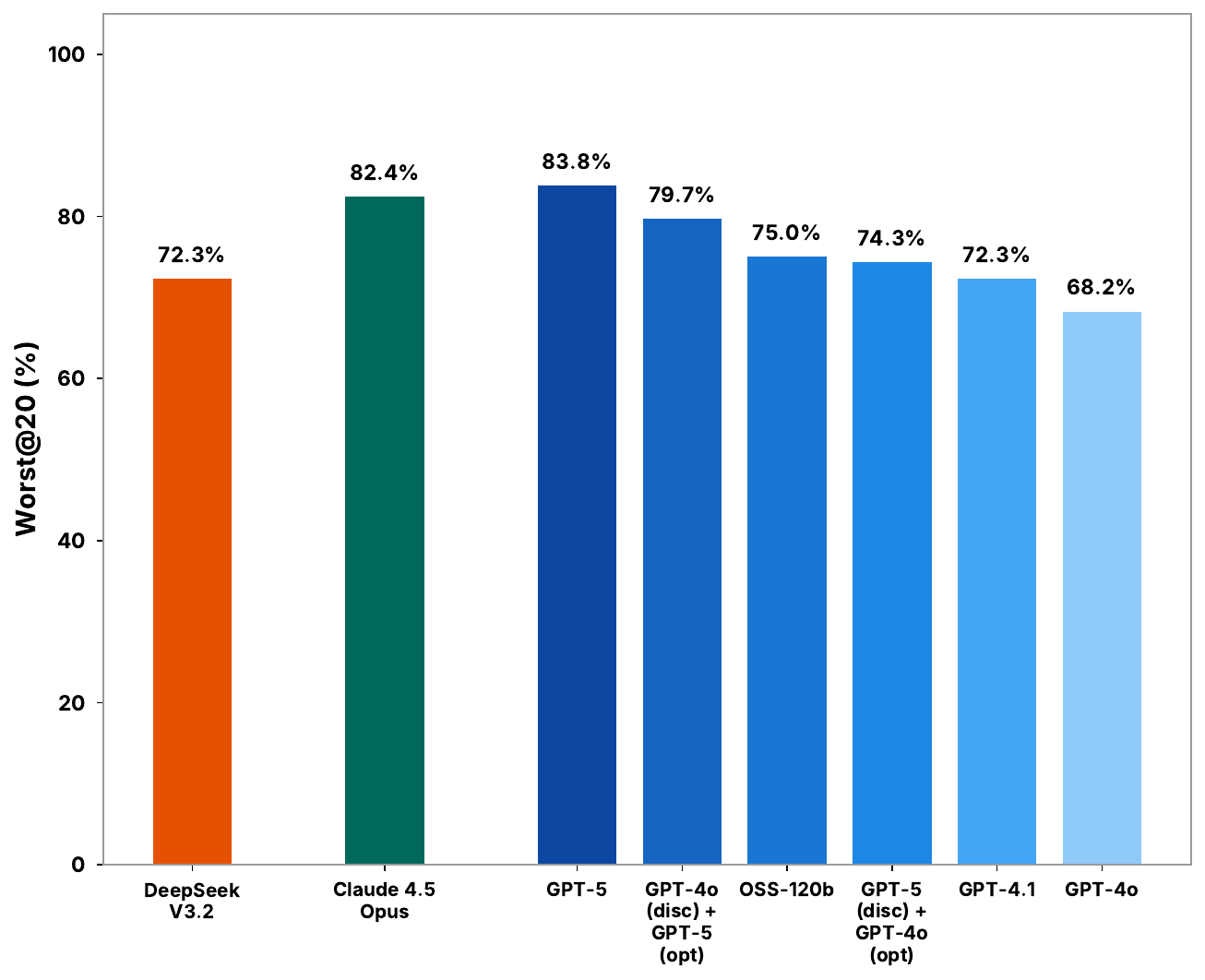}
  \caption{Each bar shows \texttt{worst@20} accuracy using different model configurations for the discriminator and optimizer. Single-model configurations use the same model for both discrimination and optimization; mixed configurations (``GPT-4o disc + GPT-5 opt'', ``GPT-5 disc + GPT-4o opt'') isolate the effect of each component. Frontier models outperform open-weight models, and optimizer capability matters more than discriminator capability.}
  \label{fig:cross_model}
\end{figure}

\Cref{fig:cross_model} shows \textsc{PrefPO} performance across different model configurations on IFEval-Hard. We test both single-model configurations (where the same model serves as the discriminator and optimizer) and mixed configurations (where discriminator and optimizer differ), all with GPT-4o as the task model. Among single-model configurations, frontier models perform best: GPT-5 achieves 83.8\% (the performance difference with other GPT-5 reported results on IFEval-Hard is due to sampling responses from OpenRouter instead of OpenAI across all models in this experiment for consistency) and Claude 4.5 Opus achieves 82.4\%. Mid-tier models show lower but reasonable performance: GPT-4.1 at 72.3\% and GPT-4o at 68.2\%. Open-weight models fall between: GPT-OSS-120b at 75.0\% and DeepSeek V3.2 at 72.3\%. Mixed configurations reveal that optimizer capability matters more than discriminator capability. Starting from GPT-4o for both components (68.2\%), upgrading only the optimizer to GPT-5 improves performance by 11.5 percentage points (to 79.7\%), while upgrading only the discriminator to GPT-5 improves performance by 6.1 percentage points (to 74.3\%). This suggests practitioners with limited compute should prioritize a stronger optimizer.

\FloatBarrier

\subsection{Component Effectiveness Testing}
\label{app:component}

To validate the algorithm's core components, we measure discriminator and optimizer performance across both tasks. We report three metrics: \textit{discriminator accuracy} (how often the discriminator selects the better-performing prompt, with ties counted as correct), \textit{optimizer accuracy} (how often the optimized prompt outperforms the input prompt), and \textit{mean improvement} (average score increase per iteration). We report these numbers with the number of examples used at discrimination time as per our scaling experiment.

\paragraph{Labeled Setting (BBH).} \Cref{tab:labeled-component} shows component effectiveness on the Disambiguation task with ground-truth labels. All three metrics scale with training examples: discriminator accuracy improves from 54.0\% (5 examples) to 90.0\% (50 examples), and optimizer accuracy from 56.7\% to 89.3\%. Both operate well above random chance with 15+ training examples, aligning with the performance gains observed at larger training splits in \Cref{fig:scaling}.

\paragraph{Label-Free Setting (BBH).} \Cref{tab:unlabeled-component} shows results without labels on the Disambiguation task. Unlike the labeled setting, component accuracy does not improve with more examples. The discriminator accuracy hovers around 50\% and optimizer accuracy around 32-47\%, with negative mean improvement throughout. This explains why unlabeled performance remains flat across data splits (\Cref{fig:scaling}): without labels, the discriminator must infer correctness from outputs alone, which is difficult when the base model achieves only 72\% on this task (see \Cref{tab:bbh_tasks_results}).

\paragraph{Label-Free Setting (IFEval-Hard).} On IFEval-Hard, which requires no labels by design, results are markedly different. Discriminator accuracy reaches 82.3\%, optimizer accuracy 84.3\%, and mean improvement 13.6\%. This demonstrates that label-free optimization is effective when the task provides clear, verifiable criteria—unlike BBH where correctness is difficult to assess without ground truth.

\begin{table}[H]
\centering
\caption{Component effectiveness with labeled training examples (Disambiguation).}
\label{tab:labeled-component}
\begin{tabular}{rccc}
\toprule
\textbf{Examples} & \textbf{Discriminator Accuracy} & \textbf{Optimizer Accuracy} & \textbf{Mean Improvement} \\
\midrule
5  & $0.540 \pm 0.121$ & $0.567 \pm 0.150$ & $0.005 \pm 0.022$ \\
10 & $0.627 \pm 0.095$ & $0.620 \pm 0.112$ & $0.025 \pm 0.018$ \\
15 & $0.760 \pm 0.144$ & $0.640 \pm 0.100$ & $0.054 \pm 0.021$ \\
20 & $0.760 \pm 0.144$ & $0.820 \pm 0.134$ & $0.091 \pm 0.024$ \\
30 & $0.887 \pm 0.060$ & $0.873 \pm 0.109$ & $\bm{0.124 \pm 0.030}$ \\
40 & $0.853 \pm 0.078$ & $0.833 \pm 0.091$ & $0.114 \pm 0.020$ \\
50 & $\bm{0.900 \pm 0.068}$ & $\bm{0.893 \pm 0.085}$ & $0.107 \pm 0.010$ \\
\bottomrule
\end{tabular}
\end{table}

\begin{table}[h]
\centering
\caption{Component effectiveness without ground-truth labels across tasks.}
\label{tab:unlabeled-component}
\small
\begin{tabular}{lcccc}
\toprule
\textbf{Task} & \textbf{Examples} & \textbf{Discriminator Accuracy} & \textbf{Optimizer Accuracy} & \textbf{Mean Improvement} \\
\midrule
Disambiguation & 5  & $0.540 \pm 0.087$ & $0.407 \pm 0.153$ & $-0.031 \pm 0.015$ \\
Disambiguation & 10 & $0.507 \pm 0.095$ & $0.467 \pm 0.094$ & $-0.029 \pm 0.017$ \\
Disambiguation & 15 & $0.487 \pm 0.090$ & $0.413 \pm 0.098$ & $-0.027 \pm 0.032$ \\
Disambiguation & 20 & $0.453 \pm 0.151$ & $0.380 \pm 0.116$ & $-0.033 \pm 0.024$ \\
Disambiguation & 30 & $0.420 \pm 0.130$ & $0.387 \pm 0.160$ & $-0.044 \pm 0.030$ \\
Disambiguation & 40 & $0.427 \pm 0.090$ & $0.320 \pm 0.154$ & $-0.046 \pm 0.036$ \\
Disambiguation & 50 & $0.447 \pm 0.137$ & $0.400 \pm 0.158$ & $-0.054 \pm 0.047$ \\
\midrule
IFEval-Hard & 1 & $0.823 \pm 0.152$ & $0.843 \pm 0.147$ & $0.136 \pm 0.166$ \\
\bottomrule
\end{tabular}
\end{table}





\end{document}